\documentclass{article}

\usepackage[preprint]{neurips_2024}

\usepackage[utf8]{inputenc} 
\usepackage[T1]{fontenc}    
\usepackage{hyperref}       
\usepackage{url}            
\usepackage{booktabs}       
\usepackage{amsfonts}       
\usepackage{nicefrac}       
\usepackage{microtype}      
\usepackage{xcolor}         

\usepackage{graphicx}
\usepackage{subfigure}
\usepackage{subcaption}
\usepackage{amsmath}
\usepackage{amssymb}
\usepackage{multirow}
\usepackage{bbm}
\usepackage{algorithm}
\usepackage{algorithmicx}
\usepackage{stmaryrd}
\usepackage[noend]{algpseudocode}

\RequirePackage{xspace}
\makeatletter
\DeclareRobustCommand\onedot{\futurelet\@let@token\@onedot}
\def\@onedot{\ifx\@let@token.\else.\null\fi\xspace}

\def\eg{\emph{e.g}\onedot}

\makeatother

\setcitestyle{numbers,square}

\def\custompurple{\textcolor[rgb]{0.5, 0, 0.5}}
\def\customyellow{\textcolor[rgb]{1.0, 0.8, 0.4}}
\def\customblue{\textcolor[rgb]{0.4, 0.6, 1.0}}

\title{Incremental Multiview Point Cloud Registration with Two-stage Candidate Retrieval}

\author{%
  Shiqi Li \qquad Jihua Zhu \qquad Yifan Xie\\
  Xi'an Jiaotong University
  \And 
  Mingchen Zhu \\
  UC Davis
}

\begin{document}

\maketitle

\begin{abstract}
Multiview point cloud registration serves as a cornerstone of various computer vision tasks. Previous approaches typically adhere to a global paradigm, where a pose graph is initially constructed followed by motion synchronization to determine the absolute pose. However, this separated approach may not fully leverage the characteristics of multiview registration and might struggle with low-overlap scenarios. In this paper, we propose an incremental multiview point cloud registration method that progressively registers all scans to a growing meta-shape. To determine the incremental ordering, we employ a two-stage coarse-to-fine strategy for point cloud candidate retrieval. The first stage involves the coarse selection of scans based on neighbor fusion-enhanced global aggregation features, while the second stage further reranks candidates through geometric-based matching. Additionally, we apply a transformation averaging technique to mitigate accumulated errors during the registration process. Finally, we utilize a Reservoir sampling-based technique to address density variance issues while reducing computational load. Comprehensive experimental results across various benchmarks validate the effectiveness and generalization of our approach. 
\end{abstract}

\section{Introduction}
Point cloud registration constitutes a pivotal challenge within the domains of computer vision and robotics. Predominantly, pairwise registration emerges as the prevalent mode in point cloud registration, having witnessed significant strides in recent years~\cite{huang2021predator,yu2021cofinet,qin2022geometric,yang2022one}. Nonetheless, a single pair of scans typically fails to encapsulate expansive scenes adequately. Consequently, the necessity arises for multiview point cloud registration, aimed at offering a comprehensive representation of the scenario by integrating data from multiple scans. The resultant complete scene reconstruction facilitates various applications including simultaneous localization and mapping (SLAM), autonomous driving, and Embodied AI. Despite the notable progress in pairwise registration, multiview point cloud registration garners comparatively less attention and presents persistent challenges in achieving satisfactory performance. 

The majority of multiview point cloud registration methodologies adhere to a global approach, initially constructing a pose graph through pairwise registration and subsequently employing motion synchronization to ascertain the pose of individual point cloud frames. Recent progress in global multiview point cloud registration encompasses the integration of more robust pairwise registration techniques~\cite{yu2023peal,jin2024multiway}, the formulation of diverse strategies for pose graph construction~\cite{wang2023robust}, and the innovation of optimization-based~\cite{huang2019learning} or learning-based~\cite{yew2021learning} motion synchronization methodologies. 

\begin{figure}
    \centering
    \includegraphics[width=\linewidth]{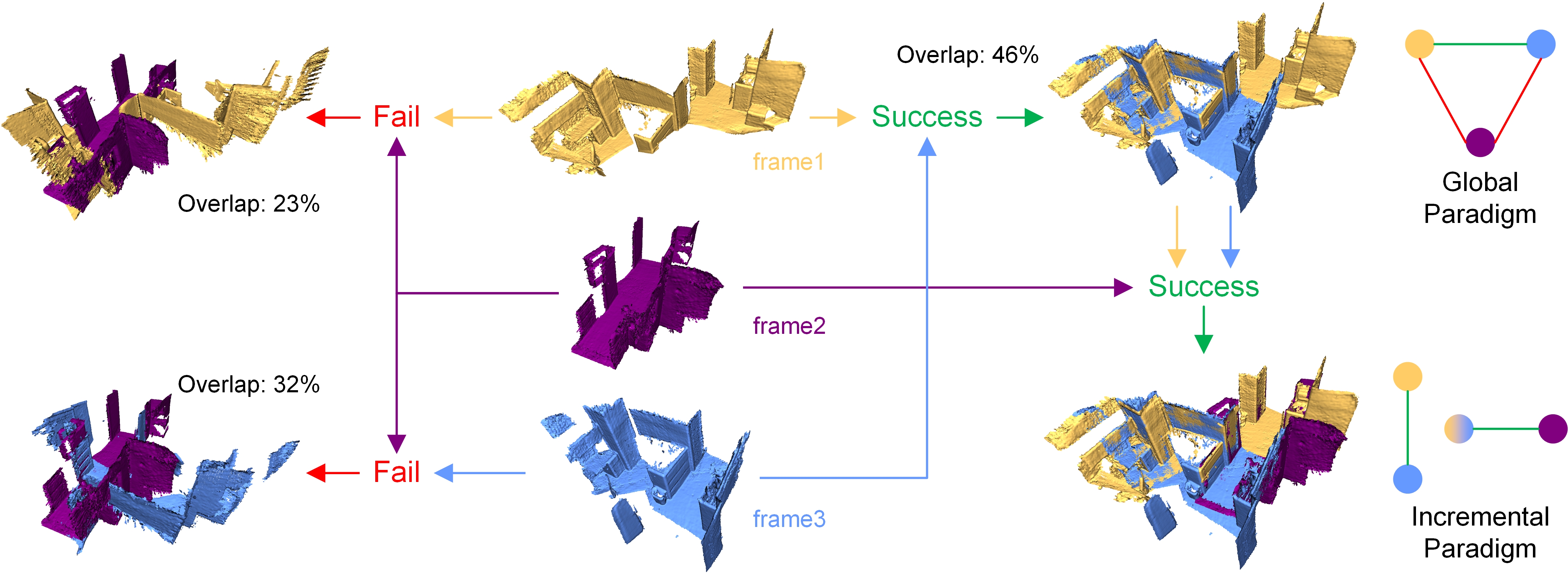}
    \caption{An example comprising \customyellow{frame1}, \custompurple{frame2}, and \customblue{frame3} within an indoor aisle. While frame1 and frame3 share some overlapping areas with frame2, these common areas predominantly consist of flat floors,  which offer limited cues for registration. However, frame1 and frame3 share sufficient structurally significant areas that enable successful alignment. The motion synchronization mechanism cannot handle the pose graph, as depicted in the top right corner, due to the absence of a correct connection to the frame2. Nevertheless, our incremental paradigm first merges frame1 and frame3 to expand the meaningful overlapping areas with frame2, ultimately achieving a complete scene, as illustrated in the bottom right corner.}
    \label{fig:example}
\end{figure}

Although these studies deliver some promising results, the global multiview registration paradigm presents inevitable drawbacks. The objective of motion synchronization is to determine a set of absolute poses $\mathbf{T}_\mathcal{V}$ that minimize consistency errors with the relative transformation $\mathbf{\tilde{T}}$ in pose graph $\mathcal{G}=\{\mathcal{V},\mathcal{E}\}$,
\begin{equation}
    \mathbf{T}_\mathcal{V} = \underset {\{\mathbf{T}_1,\mathbf{T}_2,\cdots,\mathbf{T}_{|\mathcal{V}|}\}} { \operatorname {arg\,min} }  \sum_{e_{ij}\in\mathcal{E}} d(\mathbf{\tilde{T}}_{ij}-\mathbf{T}_i\mathbf{T}_j^\top), 
    \label{eq:problem}
\end{equation}
where $d(\cdot)$ is a distance measure between two transformations. However, addressing Eq. \ref{eq:problem} may not consistently yield satisfactory outcomes. The accuracy of the estimated absolute pose depends on both the noise level of pairwise measurements and the Laplacian of the pose graph structure~\cite{boumal2014cramer}. While employing diverse graph construction and motion synchronization strategies can approximate certain lower error bounds, these bounds are significantly impacted by the accuracy and robustness of the pairwise method. For instance, as illustrated in Fig. \ref{fig:example}, if a frame within the unorganized set lacks sufficient overlapping area with all neighbors, global multiview methods may struggle to handle it.  

The incremental method represents an alternative paradigm for multiview point cloud registration. As its name suggests, incremental methods iteratively perform pairwise registration and expand the point cloud to acquire the absolute pose of each frame. In contrast to the global paradigm, the incremental method is relatively less efficient but holds the potential to yield more accurate results~\cite{gao2021incremental}. It is intuitively expected that as the point cloud iteratively grows, the overlap ratio between two frames will not be lower than that achieved through pairwise registration in the pose graph construction of the global paradigm. This characteristic may alleviate the challenges posed by low overlap issues, which are problematic for pairwise registration methods. Despite its significant potential, incremental approaches have not been fully explored; existing works primarily rely on handcrafted growth strategies~\cite{guo2014accurate} or additional prior assumptions~\cite{wu2023hierarchical}, limiting their applicability across a wide range of scenarios.

In this paper, we introduce a novel incremental multiview point cloud registration pipeline that achieves more accurate and robust estimation. Drawing inspiration from modern image retrieval~\cite{shao2023global} and visual place recognition systems~\cite{sarlin2019coarse}, we propose a two-stage coarse-to-fine point cloud retrieval module to address the critical ordering problem inherent in the incremental process. The first coarse stage is instantiated with deep learning-based global semantic features aggregation to identify potential candidate frames, while the subsequent fine stage employs geometric matching to rerank the candidates and select the best frame for point cloud expansion. Given the issue of accumulated error during the growing process, we mitigate this challenge by employing single transformation averaging. The refinement of the final pose for each point cloud is achieved by leveraging all neighbor frames that exhibit sufficient overlapping. Finally, considering the computation cost and density variance caused by the point cloud growth, we propose a Reservoir sampling-based~\cite{vitter1985random} strategy to maintain the keypoints and descriptors of the meta point cloud. We evaluate the effectiveness of our approach on 3D(Lo)Match~\cite{zeng20173dmatch,huang2021predator} and ScanNet~\cite{dai2017scannet} benchmarks. Experimental results demonstrate the superiority of our design, with our method significantly outperforming state-of-the-art methods, achieving registration recalls of 97.1\% and 87.9\% on the 3DMatch/3DLoMatch datasets, respectively.

\section{Related Work}

\paragraph{Pairwise Registration}
Pairwise point cloud registration, which forms the basis of multiview registration, can be broadly classified into two types. The first category encompasses correspondence-based approaches~\cite{choy2019fully,yu2021cofinet,huang2021predator,qin2022geometric}, which initially establish a series of tentative correspondences and then employ a robust estimator~\cite{fischler1981random,jiang2023robust,zhang20233d} or geometric hashing~\cite{hinterstoisser2016going} to determine the transformation. Notably, the Iterative Closest Point (ICP)~\cite{besl1992method} series, as representative methods, establish correspondences based on various distances in Euclidean space, albeit they are susceptible to initialization and outliers. To enhance robustness, handcrafted features have been proposed to characterize the geometric nature of point clouds in a local area~\cite{rusu2009fast}, and correspondences are established through feature matching. Recent advancements in deep learning tend to replace handcrafted features with learnable descriptors~\cite{choy2019fully,wang2022you}. Metric learning-guided neural networks automatically generate similar descriptors for consistent areas and distinct descriptors for non-overlapping areas, with cross-attention mechanisms typically applied to facilitate explicit information exchange~\cite{huang2021predator,qin2022geometric,yu2023peal}. The second category comprises correspondence-free methods~\cite{xu2022finet,jiang2024se}, which aim to directly regress the transformation from the point cloud pair instead of establishing correspondences. Although these methods have shown promising results on synthetic object-level shapes, they struggle with large-scale real-world scenes. In this study, we utilize off-the-shelf learnable correspondence-based pairwise registration methods as a plug-and-play module in our multiview registration pipeline.

\paragraph{Multiview Registration}
Multiview point cloud registration endeavors to reconstruct a holistic scene from a set of unorganized, partially overlapping point cloud scans. Analogous to the taxonomy of structure-from-motion (SfM), multiview registration can be classified into global and incremental methods. Global methods initially collect relative transformations to construct a pose graph, followed by utilizing global cycle consistency to optimize absolute poses~\cite{arrigoni2016global,arrigoni2018robust}. Recognizing that fully connected graphs may encompass numerous outliers, a pruning strategy is introduced to eliminate low-quality registration pairs~\cite{gojcic2020learning}. Additionally, a more efficient approach in~\cite{wang2023robust} directly builds a sparse graph. Motion synchronization aims to recover global poses with minimal consistency error, as mentioned in Eq. \ref{eq:problem}. Iteratively reweighted least-squares (IRLS) based schemes are prevalent in motion averaging~\cite{arrigoni2016spectral,huang2019learning,gojcic2020learning}, gradually reducing the weights of outlier edges and assigning higher weights to inlier edges by optimizing a robust loss function~\cite{chatterjee2017robust}. However, IRLS may be susceptible to local minima when undesired robust loss or reweighting schemes are employed. Beyond IRLS, recent approaches address the synchronization problem in a data-driven manner~\cite{yew2021learning,li2022rago}, often projecting transformations into a high-dimensional latent space and utilizing graph neural networks (GNN) to facilitate information interaction among the pose graph. In contrast, incremental methods iteratively merge a new point cloud frame into the growing meta-point cloud until all scans are incorporated into the final scene point cloud~\cite{guo2014accurate}. The incremental paradigm finds more popularity in the fields of geoscience and remote sensing~\cite{dong2018hierarchical,wu2023hierarchical}, particularly given the extensive overlap typically present in terrestrial laser scans between successive frames.

\section{Method}

\begin{figure}[t]
    \centering
    \includegraphics[width=\linewidth]{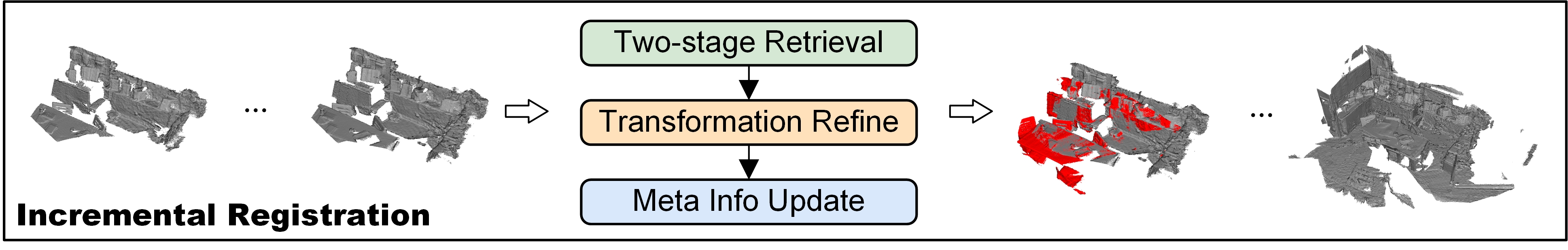}
    \caption{Overview of our pipeline: Point cloud frames are progressively incorporated into the same coordinate system, referred to as the meta-shape in this study. For each frame, the process entails two-stage frame retrieval, transformation refinement, and meta-information update.}
    \label{fig:overall}
\end{figure}


Given a set of unorganized partial overlap point cloud scans $\mathcal{P}=\{P_i|i=1,2,\dots,N\}$. The multiview registration aims to seek an absolute pose $\mathbf{T}_i=(\mathbf{R}_i, \mathbf{t}_i)\in SE(3)$ for each point cloud frame $P_i$ that correctly recovers the complete scene in a unified coordinate system. In this study, we employ local descriptor-based methods~\cite{choy2019fully,wang2022you} to align point cloud pairs. Consequently, for each point cloud $P_i$, we can extract a set of keypoints along with their associated local descriptors $\mathcal{F}_i$. The transformation is estimated based on the correspondences established through descriptor matching.

The overview of our proposed method is illustrated in Fig. \ref{fig:overall}.

\subsection{Coarse-to-Fine Frame Selection}
\label{sec:selection}

\begin{figure}
    \centering
    \includegraphics[width=\linewidth]{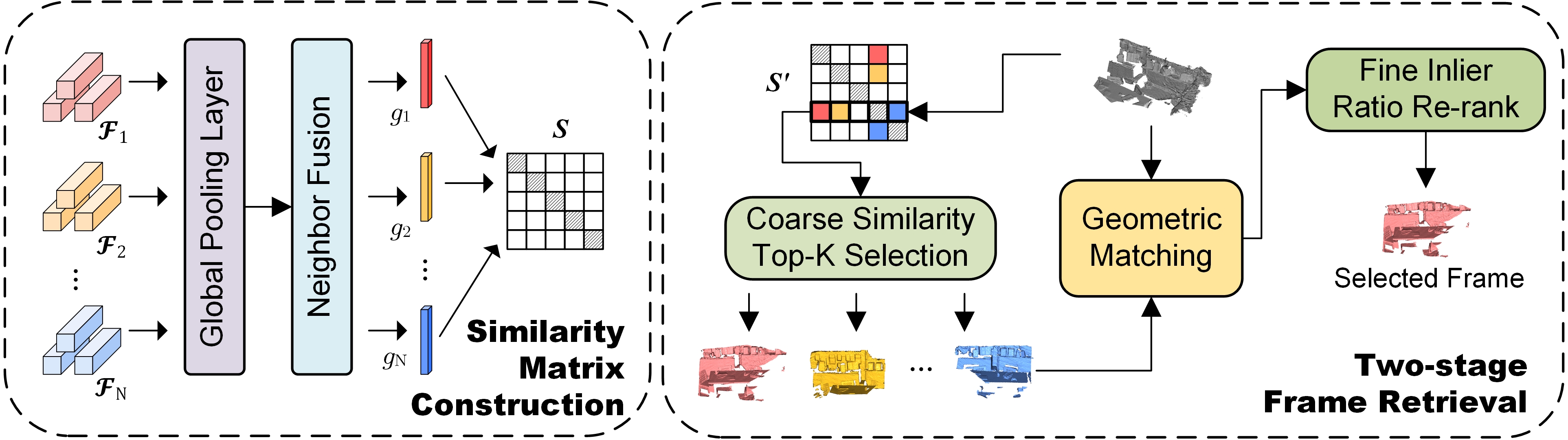}
    \caption{Left: Illustration depicting the construction of the similarity matrix. Right: Workflow of the proposed two-stage frame retrieval process.}
    \label{fig:selection}
\end{figure}

Determining the order of registration sequences is crucial for our incremental method. We adopt a two-stage coarse-to-fine approach to accurately select the newly added point cloud frame. In the first stage, point clouds similar to the meta-shape are coarsely retrieved using global features aggregated by a deep learning method. Then, the selected candidates undergo further reranking through a second stage based on geometric matching. Subsequently, the best matching frame is registered into the meta-shape, as shown in Fig. \ref{fig:selection}.

Our first stage is instantiated with a global pooling layer (\eg NetVLAD~\cite{arandjelovic2016netvlad}, GeM~\cite{radenovic2018fine}). This layer is responsible for aggregating a global feature for each point cloud frame from the associated local descriptor set $\mathcal{F}_i$,
\begin{equation}
    \Tilde{g_i}=Pooling(\mathcal{F}_i).
\end{equation}

To enhance the effectiveness of the vanilla global pooling features, we further introduce a fusion operation~\cite{shao2023global}. Formally, each global feature is refined using its neighboring features, 
\begin{equation}
    g_i=\frac{\Tilde{g_i}+\sum_{j\in\mathcal{M}_i}\beta(\Tilde{g_j}\cdot \Tilde{g_i})\Tilde{g_i}}{1+\sum_{j\in\mathcal{M}_i}\beta(\Tilde{g_j}\cdot \Tilde{g_i})},
    \label{eq:fusion}
\end{equation}
where $\mathcal{M}_i$ represents indexes of the nearest neighbors of $\Tilde{g_i}$. This fusion operation facilitates the generation of similar global features for overlapping point clouds.

We construct the similarity matrix $S\in\mathbb{R}^{N\times N}$ by mapping the distances between any two global features to $[0, 1]$,
\begin{equation}
    s_{ij}=\frac{2-\Vert g_i-g_j\Vert_2}{2}.
\end{equation}

With the matrix $S$, we initialize the meta-shape as the frame that is most similar to the others, 
\begin{equation}
    x=\underset {1\leq i \leq N} { \operatorname {arg\,max} } \sum_{j=1}^{N}s_{ij},
\end{equation}
The keypoints and descriptors of point cloud $x$ are utilized to initialize the meta-shape, while the pose of point cloud $x$ is set to the identity matrix $\mathbf{I}$.

After determining the initial meta-shape, we select the top-$k$ similar point cloud frames to the meta-shape based on their scores in $S$ as candidate frames. The procedure outlined here serves as an example using the initial meta-shape; however, the same approach is employed for arbitrary meta-shapes during the incremental process. Additionally, instead of employing the pooling layer to aggregate a new global descriptor at each step, the global similarity matrix $S$ is dynamically updated throughout the incremental registration process. Details of the global similarity matrix update will be provided in Sec. \ref{sec:update}.

Recognizing that the global feature-based retrieval in the first stage may inadvertently lose some subtle patterns during the aggregation process, we further introduce a second stage based on geometric matching to rerank the candidates. Specifically, for each candidate frame, we establish a coarse correspondence set $\mathcal{C}$ with the meta-shape using descriptor matching. Subsequently, a robust estimator (\eg RANSAC) is employed to estimate the transformation $\mathbf{T}$ between the meta-shape and the candidate frame. Finally, we compute the inlier count (IC) for each candidate frame, and the frame with the highest IC is selected to be merged into the meta-shape. IC measures the number of correspondences that confirm the estimated transformation, where a high IC value supports reliable registration. The IC can be calculated as: 
\begin{equation}
    \text{IC}=\sum_{(p, q)\in\mathcal{C}}\llbracket \Vert \mathbf{T}(p)-q\Vert_2 < \tau \rrbracket,
\end{equation}
where $\llbracket\cdot\rrbracket$ is Iverson bracket, $\tau$ is a distance threshold.

\subsection{Transformation Refinement}

The selection process identifies the most suitable frame and provides an estimated transformation $\mathbf{T}$. However, we do not solely rely on the transformation from the robust estimator; instead, we further refine it with single translation averaging. For each frame in the meta-shape, we calculate the overlap ratio with the selected frame using: 

\begin{equation}
    o_{ij}=\frac{\sum_{p\in P_i}\llbracket \Vert \mathbf{T}(p)-\text{NN}(\mathbf{T}(p),P_j)\Vert_2 < \tau \rrbracket+\sum_{q\in P_j}\llbracket \Vert q-\text{NN}(q,\mathbf{T}(P_i))\Vert_2 < \tau \rrbracket}{|P_i|+|P_j|},
    \label{eq:overlap}
\end{equation}
where $\text{NN}(\cdot)$ denotes the spatial nearest neighbor.

For frames that contain more than 30\% overlap with the selected frame, we solve the Orthogonal Procrustes problem to compute a new transformation. If the selected frame yields more than one new transformation, we apply a single transformation averaging method to average them. Otherwise, the transformation obtained from the robust estimator is directly employed. 

The transformation averaging process comprises rotation averaging and translation averaging. For rotation averaging, we utilize a simplified single rotation averaging method from~\cite{lee2020robust} to compute the averaged rotation matrix. At the beginning, the rotation $\mathbf{R}$ from the robust estimator is set as the initial averaging estimation. Subsequently, both $\mathbf{R}$ and the rotations $\hat{\mathbf{R}}$ from the newly observed transformations are converted into vectors. Next, we gather the residual vectors $v$ between $\mathbf{R}$ and $\hat{\mathbf{R}}$ in vector form and calculate associated norms. The averaging estimation is then updated to a weighted sum of residual vectors, where the weights correspond to the overlapping ratio in Eq. \ref{eq:overlap}. This procedure is iterated until the averaging estimation converges or reaches the maximum iteration rounds. Finally, the averaging estimation vector is converted into matrix form and projected onto the $SO(3)$ space using SVD to obtain the final $\Bar{\mathbf{R}}$. A detailed algorithm can be found in the Appendix \ref{app:averaging}.

Based on the rotation averaging result, the translation averaging can be formulated as follows:
\begin{equation}
    \Bar{\mathbf{t}}=\underset {\mathbf{t}\in\mathbb{R}^{3\times 1}} { \operatorname {arg\,min} } \sum_{i=1}^{M} w_i\Vert\hat{\mathbf{t}}_i-\hat{\mathbf{R}}_i\Bar{\mathbf{R}}^\top\mathbf{t} \Vert_2,
\end{equation}
where $M$ the number of newly observed transformation, and $w_i$ is the associated overlapping ratio. This problem can be solved using the least squares method. We construct three zero block matrices: $\mathbf{A}\in\mathbb{R}^{3M\times 3}$, $\mathbf{B}\in\mathbb{R}^{3M\times 1}$, and $\mathbf{W}\in\mathbb{R}^{3M\times 3M}$, where $\mathbf{A}$ and $\mathbf{W}$ consist of $3\times 3$ blocks, and B consists of $3\times 1$ blocks. For each newly observed transformation, the $i$-th block on the diagonal of $\mathbf{W}$ is set to $w_i\mathbf{I}$, while the $i$-th blocks in $\mathbf{A}$ and $\mathbf{B}$ are set to $\hat{\mathbf{R}}_i\Bar{\mathbf{R}}^\top$ and $\hat{\mathbf{t}}_i$, respectively. The final averaged translation can then be calculated as:
\begin{equation}
    \Bar{\mathbf{t}}=(\mathbf{A}^\top\mathbf{W}\mathbf{A})^{-1}\mathbf{A}^\top\mathbf{W}\mathbf{B}.
\end{equation}

The result of the transformation averaging $\Bar{\mathbf{T}}=(\Bar{\mathbf{R}},\Bar{\mathbf{t}})$ is utilized to determine the pose of the selected frame. It is worth noting that to be consistent with the coordinate system definition commonly used in the registration field the final absolute pose is the inverse of the calculated transformation.

\subsection{Meta Information Update}
\label{sec:update}

\begin{figure}
    \centering
    \includegraphics[width=\linewidth]{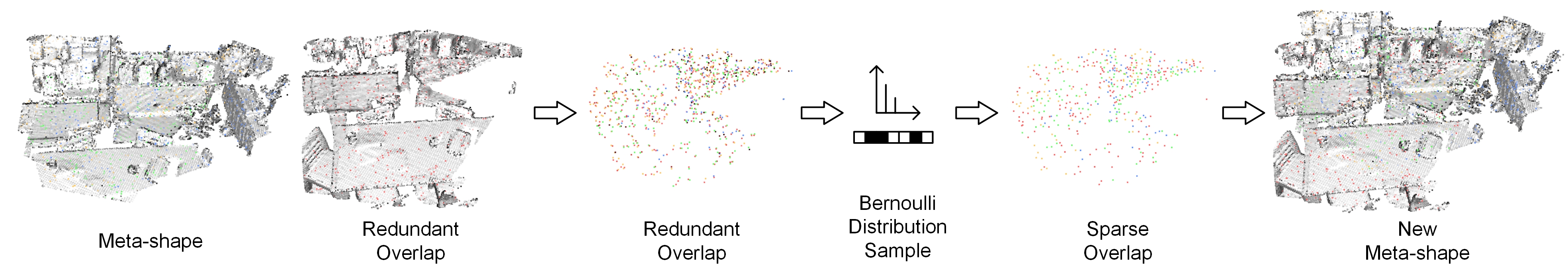}
    \caption{An illustration of the meta information update process.}
    \label{fig:sample}
\end{figure}

After obtaining the pose of the selected frame, we utilize it to update the meta-shape. The simplest method involves directly adding the keypoints and descriptors of the selected frame to the meta-shape. However, this straightforward merging operation introduces numerous redundant points and descriptors in the overlapping areas, resulting in two drawbacks. Firstly, it increases memory usage and computational costs. Secondly, this approach causes density variance in the meta-shape, which adversely affects feature matching and registration processes.

To mitigate redundancy, we employ a sampling strategy to update the keypoints and descriptors, as illustrated in Fig. \ref{fig:sample}. Initially, we use the estimated pose to align the selected point cloud frame with the coordinate system of the meta-shape. Subsequently, we identify mutual nearest point pairs in Euclidean space between the meta-shape and the selected frame, considering pairs with distances below the threshold $\tau$ as redundant. Next, we randomly sample $p_{k_1,k_2,\dots,k_n}$ from a multivariate Bernoulli distribution $P\{X_1=k_1, X_2=k_2,\dots,X_n=k_n\}$ to determine the source of the keypoints and descriptors in the overlapping area. Here, the number of variables $n$ equals the count of redundant pairs, with each variable $X_i$ following an independent Bernoulli distribution whose probability mass function can be expressed as (derivation in Appendix \ref{app:pmf}):
\begin{equation}
    P(X_i=k)=\frac{r_i^{1-k}}{r_i+1},k\in\{0,1\},
\end{equation}
where $r_i$ represents the redundancy count for point in pair $i$. This Reservoir sampling-based~\cite{vitter1985random} design ensures that overlapping points covered by multiple point cloud frames are uniformly sampled, thus yielding a high-quality meta-shape for multiview registration. For each redundant pair, if the corresponding variable equals 0, the keypoint and descriptor from the meta-shape are used; otherwise, those from the selected frame are utilized. The points and descriptors from the sampled redundant pairs are then combined with other non-pairing ones to form the new meta-shape.

In addition to updating the keypoints and descriptors, the global similarity matrix $S$ is also updated. We utilize the row corresponding to the selected frame to update the row of the meta-shape. The similarity is set to the maximum value between the two rows, except the frame already present in the meta-shape, which is set to 0.

Ultimately, the updated meta-shape and matrix $S$ revert to the two-stage selection process outlined in Sec. \ref{sec:selection} to incorporate a new frame until all frames are included in the meta-shape.

\subsection{Loss}
To train the coarse stage in our selection module, we employ the smooth L1 loss~\cite{girshick2015fast} between the predicted similarity score $s_{ij}$ and the ground truth overlapping ratio. 

\section{Experiments}

\subsection{Dataset}

The 3DMatch~\cite{zeng20173dmatch} is an indoor dataset commonly utilized for registration tasks. It comprises 46 training scenes, 8 validation scenes, and 8 testing scenes. Our evaluation adheres to the standard practice of assessing both the 3DMatch and 3DLoMatch~\cite{huang2021predator} test splits. Notably, the overlap of point cloud pairs in 3DMatch exceeds 30\%, whereas in 3DLoMatch, it ranges between 10\% and 30\%.

The ScanNet~\cite{dai2017scannet} dataset comprises RGB-D data collected from over 1500 indoor scenes. Following the operation outlined in~\cite{gojcic2020learning,wang2023robust}, we evaluate a subset of 32 scenes, consisting of 960 scans and a total of 13920 pairs. To generate point clouds, we randomly select 30 RGB-D images from each scene, spaced 20 frames apart, and convert them into point clouds. Notably, the temporal sequence of frames is disregarded during this process. The substantial temporal gap between frames, coupled with the omission of temporal sequence, renders the test setting exceptionally challenging.

\subsection{Baseline and Protocol}
We compare our method with EIGSE3~\cite{arrigoni2016spectral}, L1-IRLS~\cite{chatterjee2017robust}, RotAvg~\cite{chatterjee2017robust}, LMVR~\cite{gojcic2020learning}, LITS~\cite{yew2021learning}, HARA~\cite{lee2022hara}, and SGHR~\cite{wang2023robust}. EIGSE3 employs spectral decomposition for motion synchronization and integrates IRLS to enhance robustness. L1-IRLS and RotAvg are classical rotation averaging methods utilizing IRLS with $\ell_1$ and $\ell_{1/2}$ loss functions, respectively. LMVR is an end-to-end learning approach that combines pairwise registration and motion synchronization within a single network. LITS is a motion synchronization method leveraging GNN for learning. HARA utilizes a cycle consistency-based spanning tree to reject outliers in rotation averaging. SGHR is a multiview registration method based on sparse pose graphs and IRLS transformation synchronization. We evaluate the multiview performance of all methods using the FCGF~\cite{choy2019fully} and YOHO~\cite{wang2022you} as pairwise registration methods, with the exception of LMVR, which includes its own pairwise module.

For global paradigm methods, we offer three types of pose graphs: full, pruned, and sparse. "Full" denotes the exhaustive registration of all scan pairs. "Pruned" entails retaining only those scan pairs where the median point distance in the registered overlapping region is less than 0.05m from the fully connected graph~\cite{gojcic2020learning}. "Sparse" involves selecting pairs based solely on the technique outlined in~\cite{wang2023robust}.

\subsection{Metrics}
Following the protocol in~\cite{gojcic2020learning,wang2023robust}, we assess multiview registration by utilizing relative transformations calculated from the recovered absolute poses. 

For the 3D(Lo)Match datasets, we utilize the Registration Recall (RR) metric. RR measures the fraction of point cloud pairs with a transformation error smaller than a specified threshold (0.2m). 

For ScanNet, we present the mean, median, and empirical cumulative distribution functions (eCDF) of the rotation error $re$ and translation error $te$. The $re$ and $te$ are defined as:
\begin{equation}
    re=\arccos\left(\frac{tr(\mathbf{R}^{pred\top}\mathbf{R}^{gt})-1}{2}\right), te=\Vert \mathbf{t}^{pred}-\mathbf{t}^{gt}\Vert_2.
\end{equation}

\subsection{Results}
\begin{figure}[H]
    \centering
    \subfigure[HARA]{
    \begin{minipage}[b]{0.23\linewidth}
    \includegraphics[width=1\linewidth]{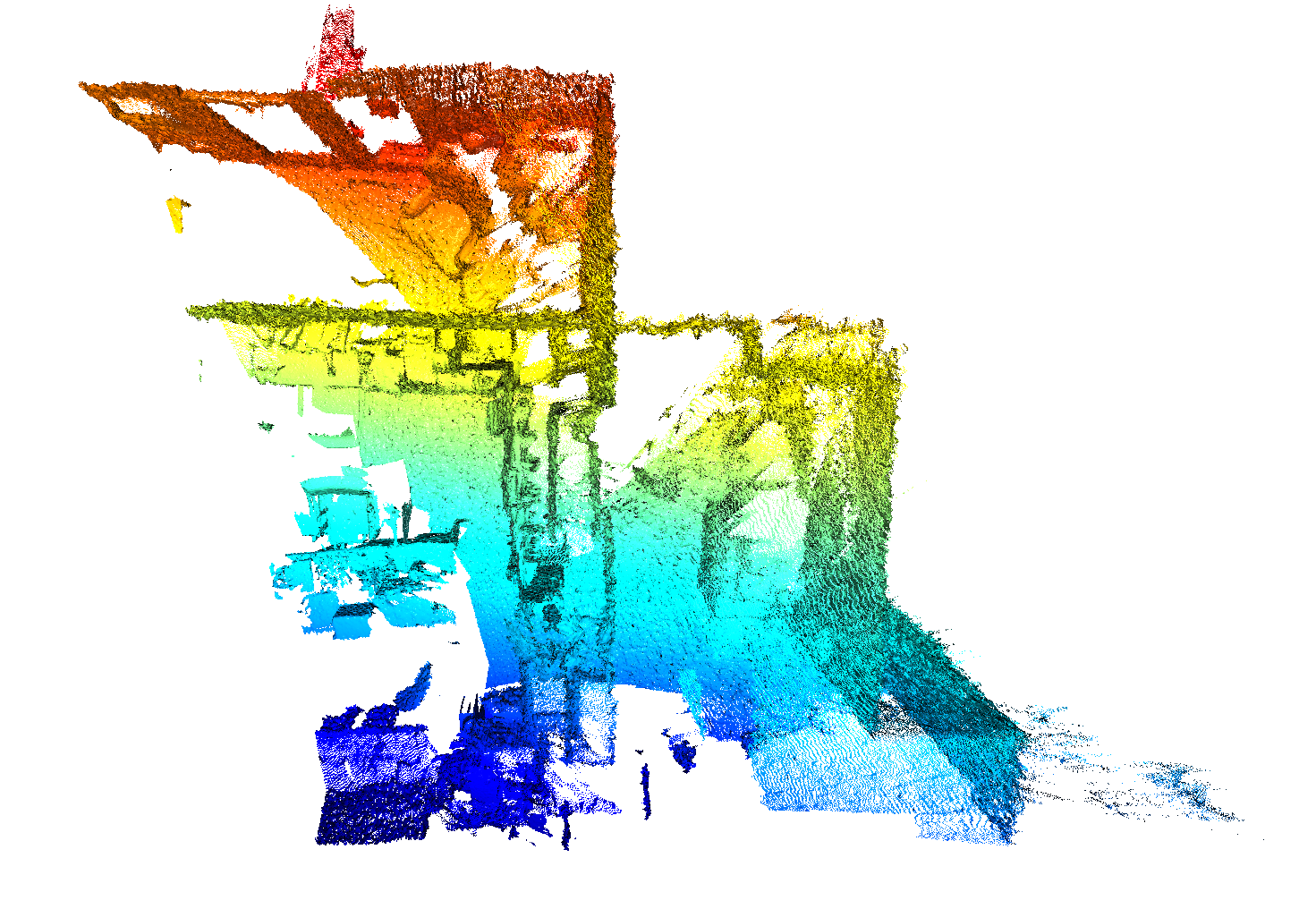}
    \includegraphics[width=1\linewidth]{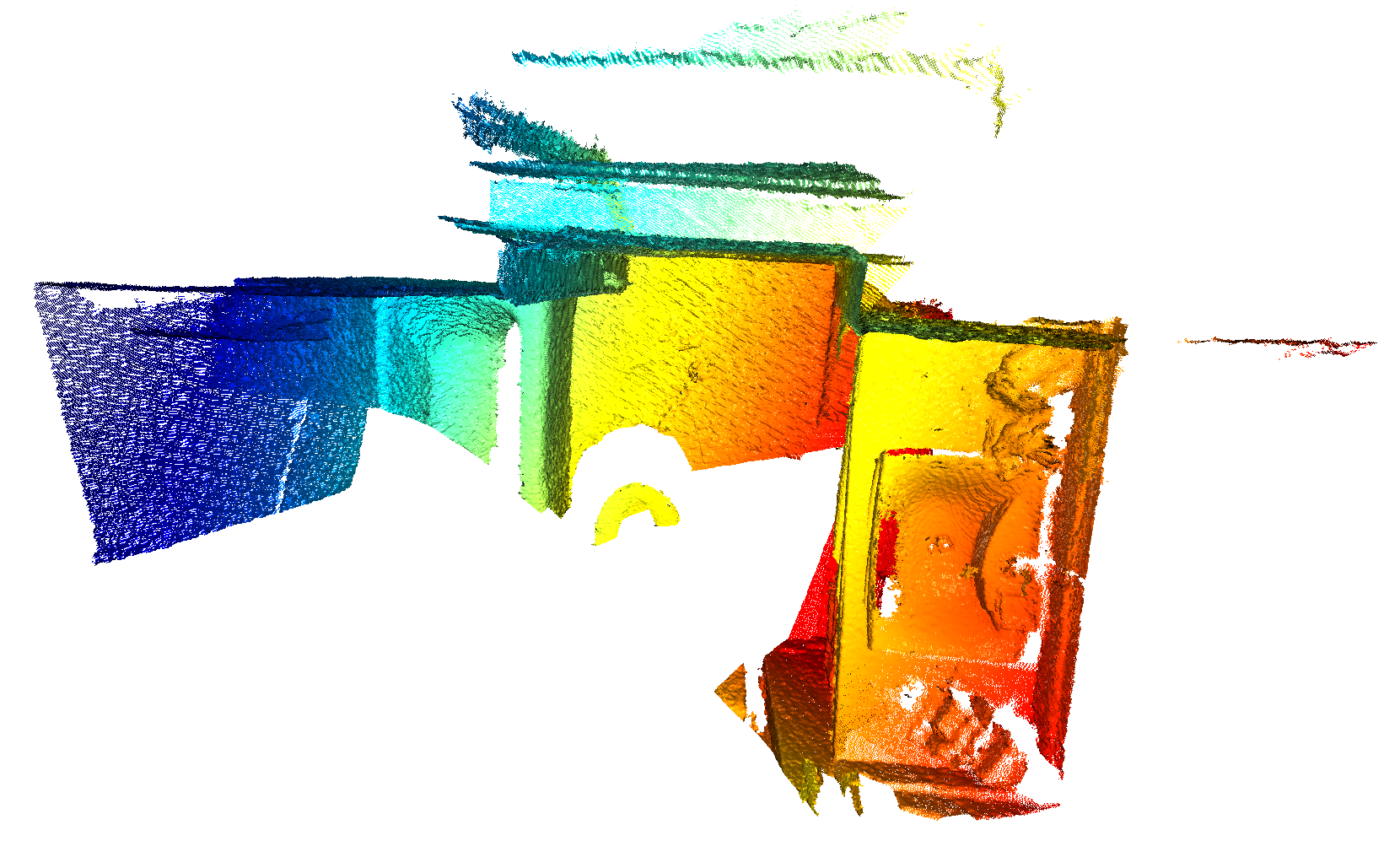}
    \includegraphics[width=1\linewidth]{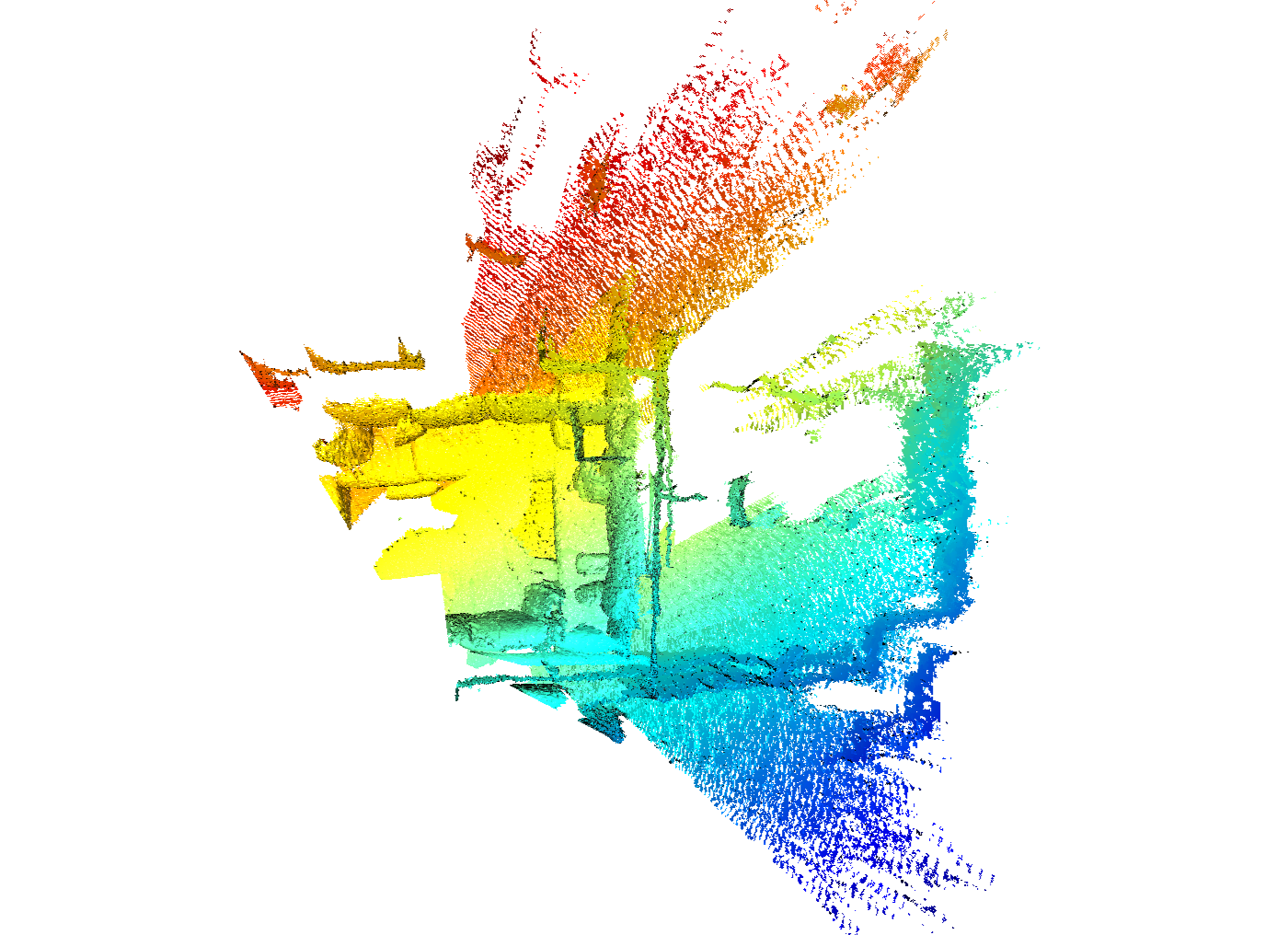}
    \end{minipage}}
    \subfigure[SGHR]{
    \begin{minipage}[b]{0.23\linewidth}
    \includegraphics[width=1\linewidth]{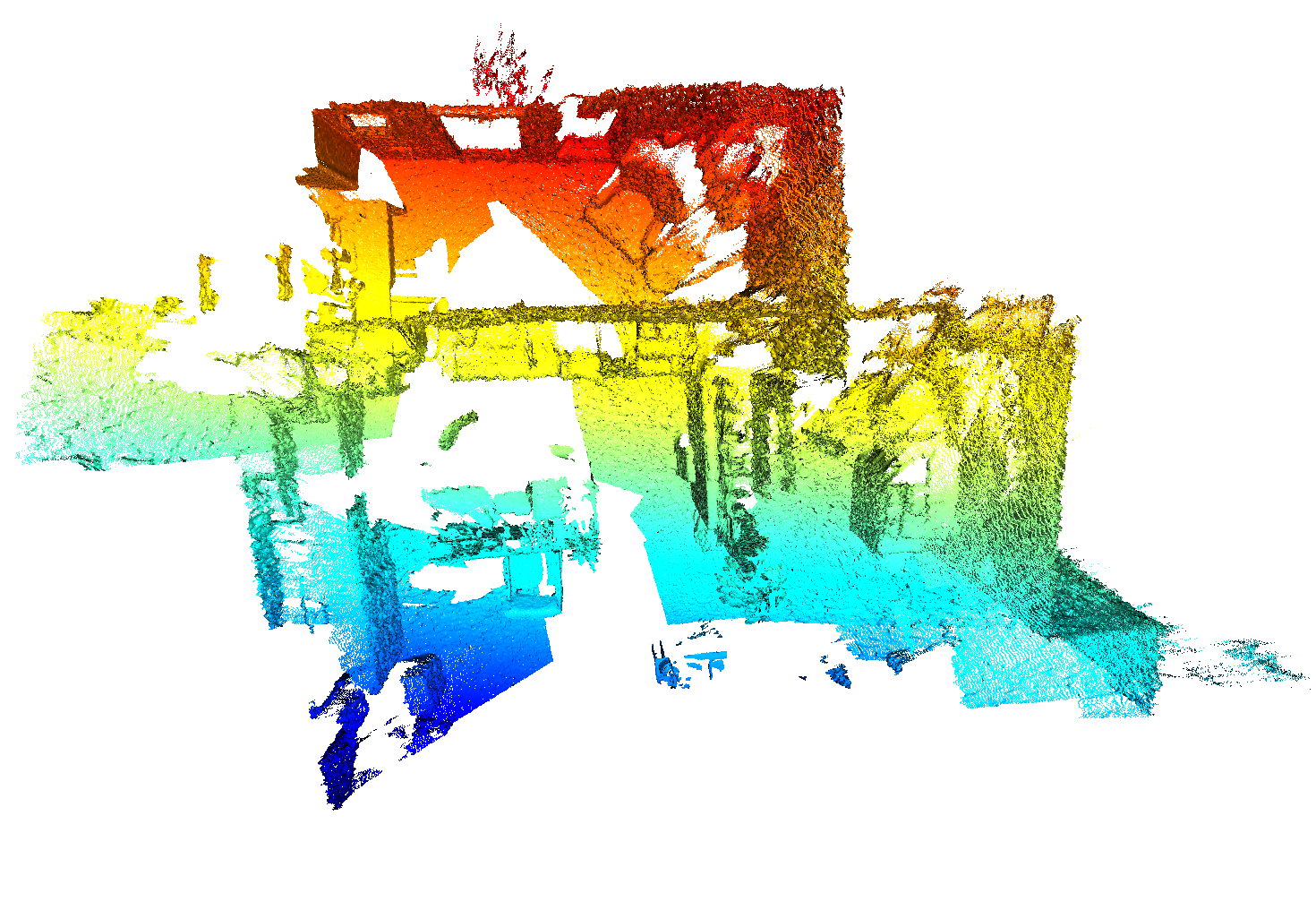}
    \includegraphics[width=1\linewidth]{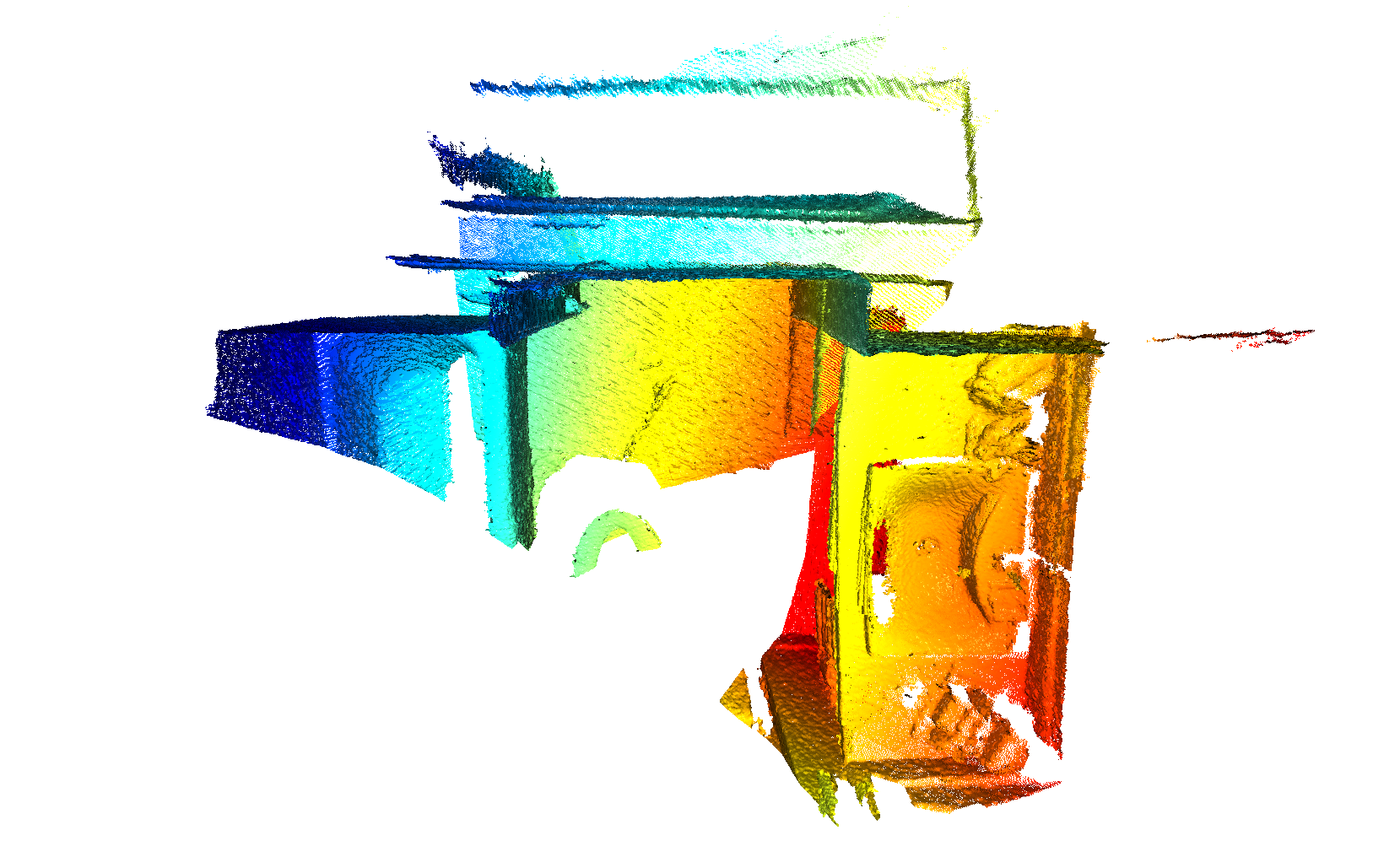}
    \includegraphics[width=1\linewidth]{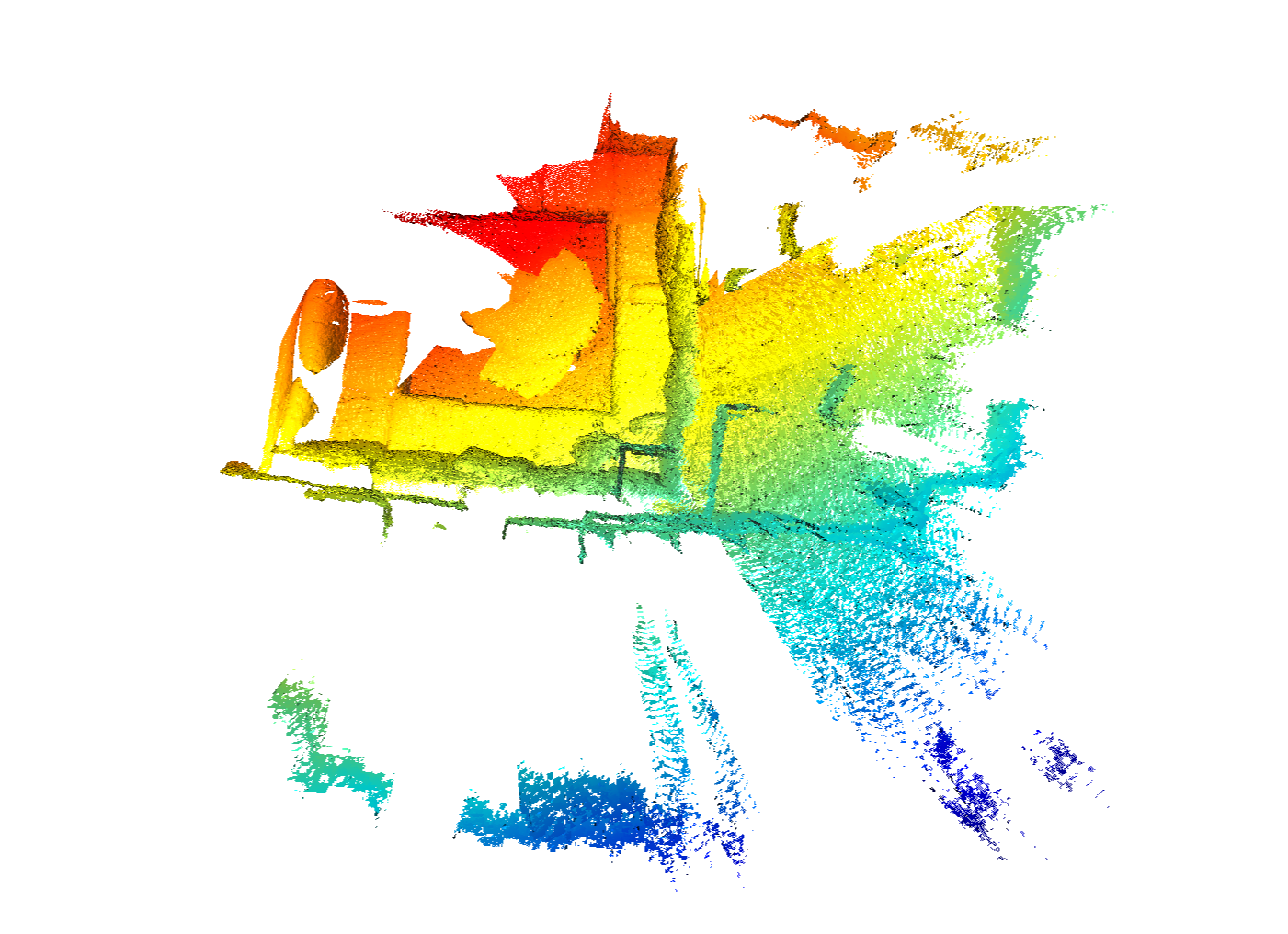}
    \end{minipage}}
    \subfigure[Ours]{
    \begin{minipage}[b]{0.23\linewidth}
    \includegraphics[width=1\linewidth]{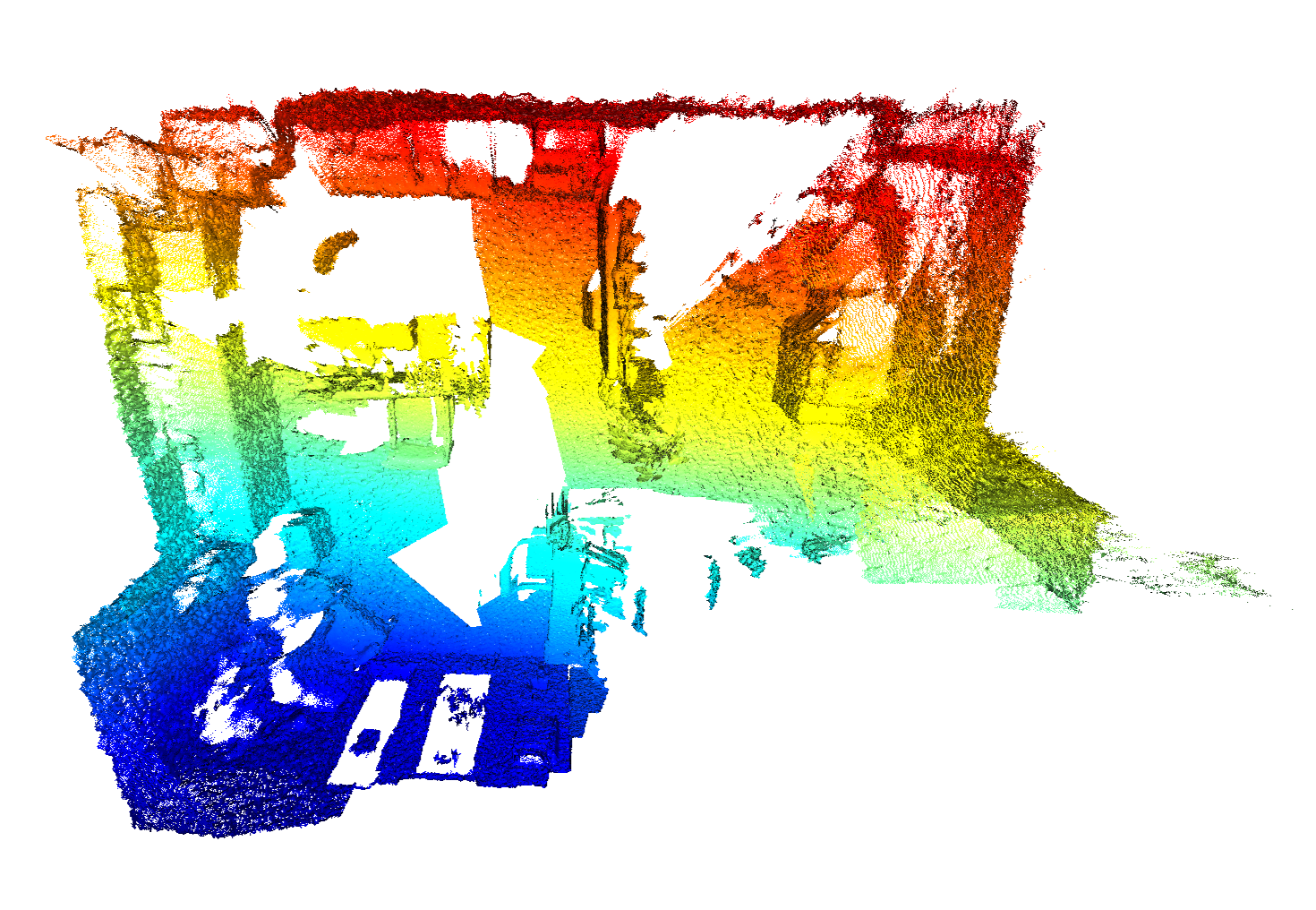}
    \includegraphics[width=1\linewidth]{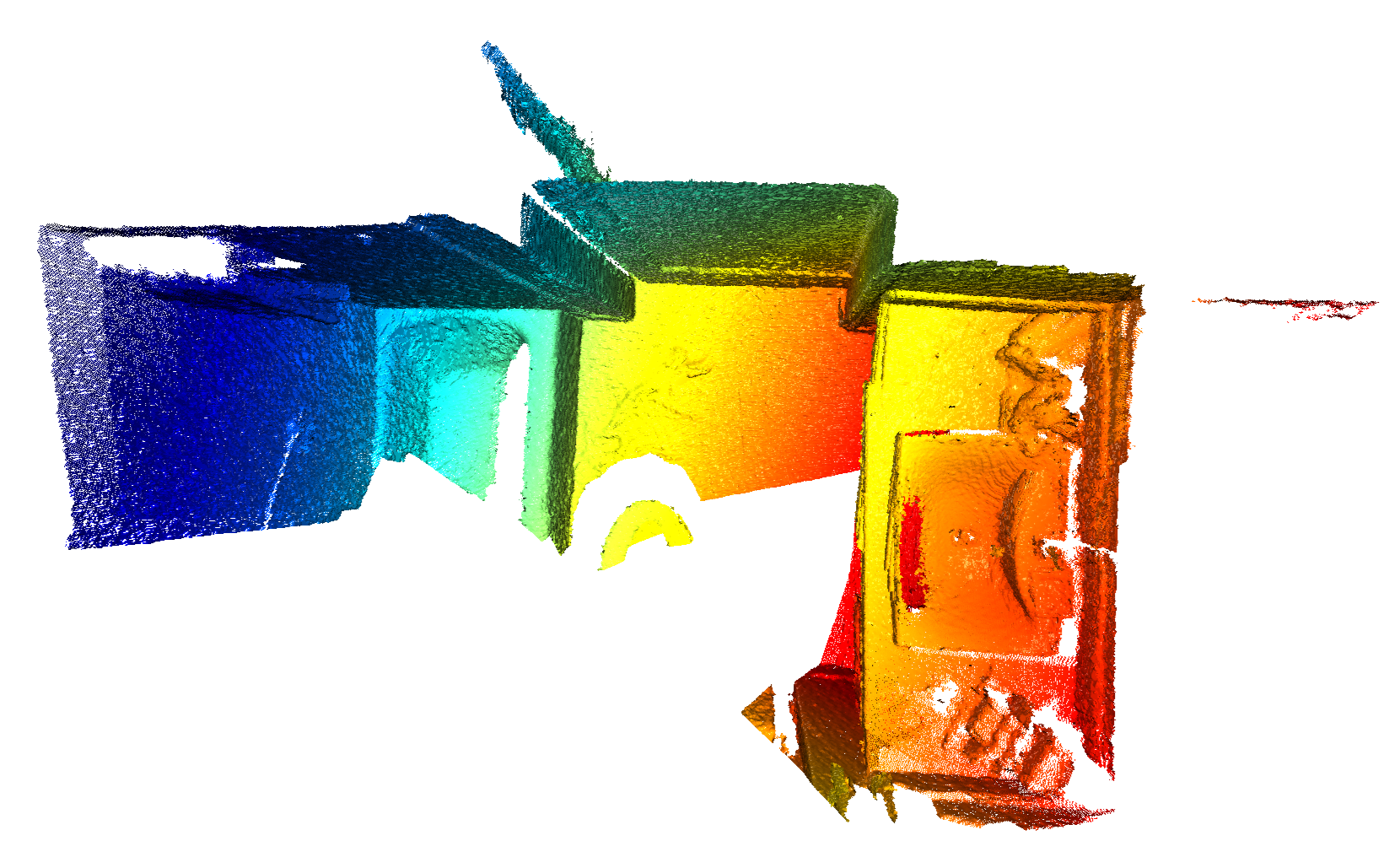}
    \includegraphics[width=1\linewidth]{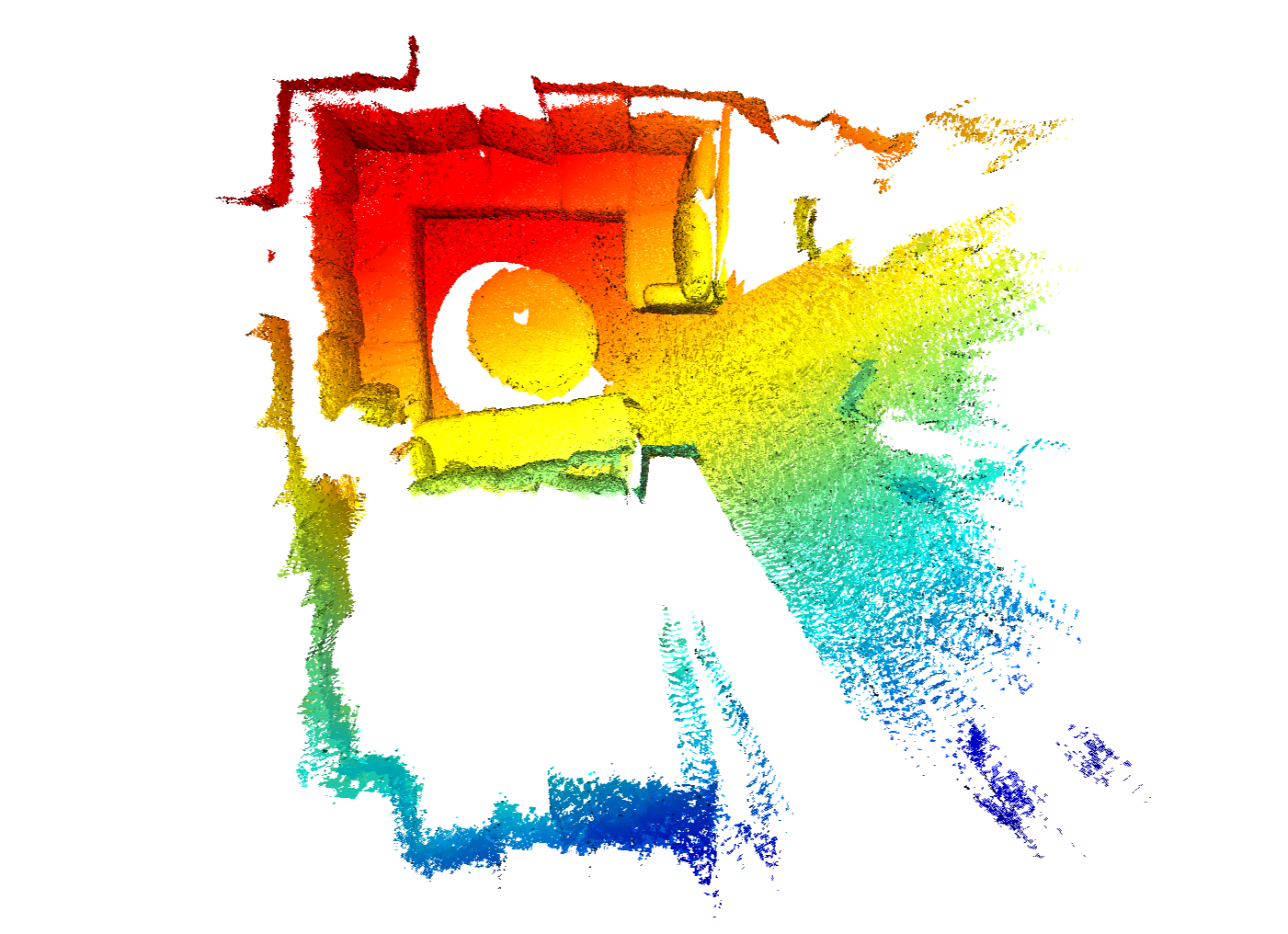}
    \end{minipage}}
    \subfigure[Ground Truth]{
    \begin{minipage}[b]{0.23\linewidth}
    \includegraphics[width=1\linewidth]{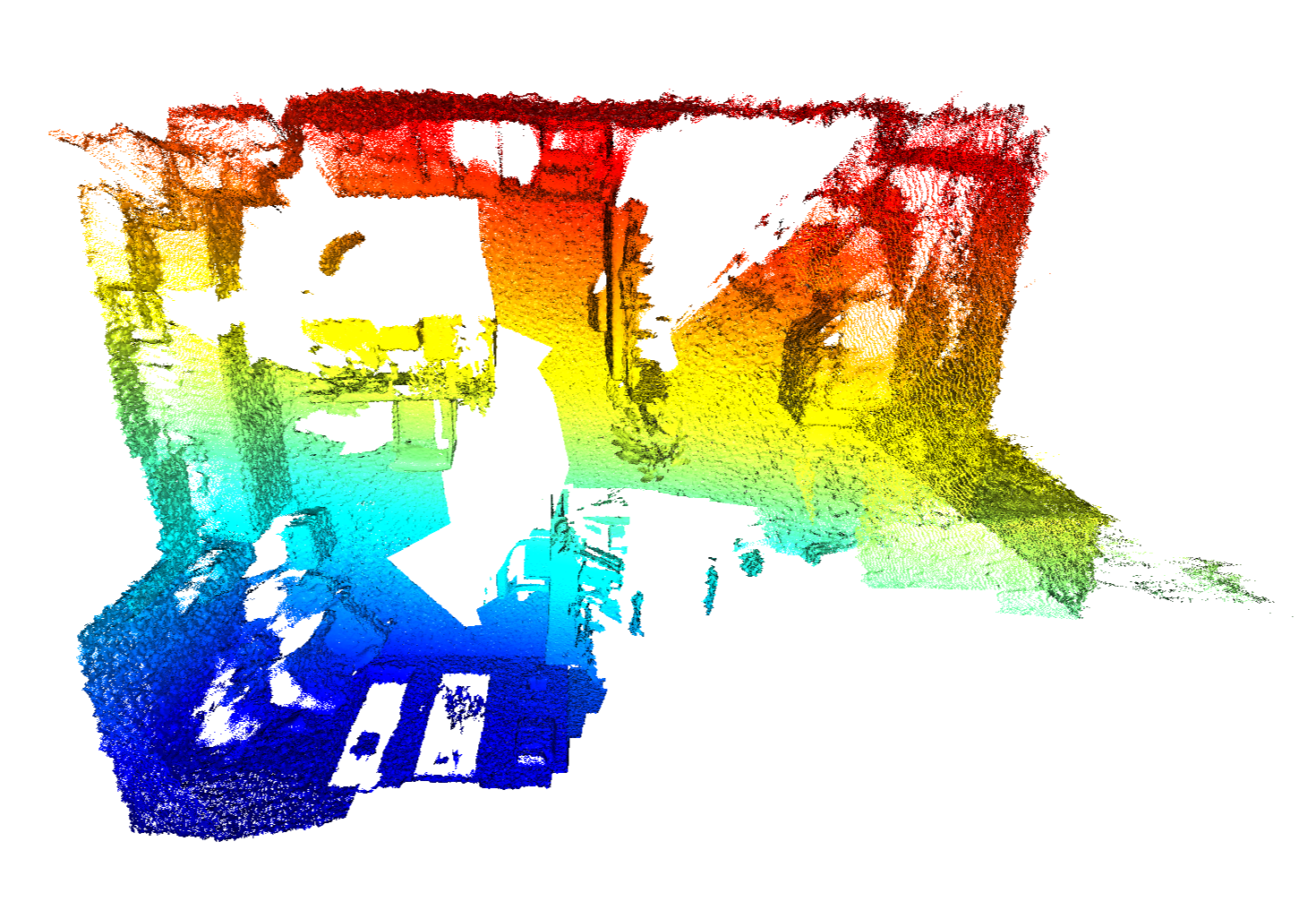}
    \includegraphics[width=1\linewidth]{figures/676_ours.png}
    \includegraphics[width=1\linewidth]{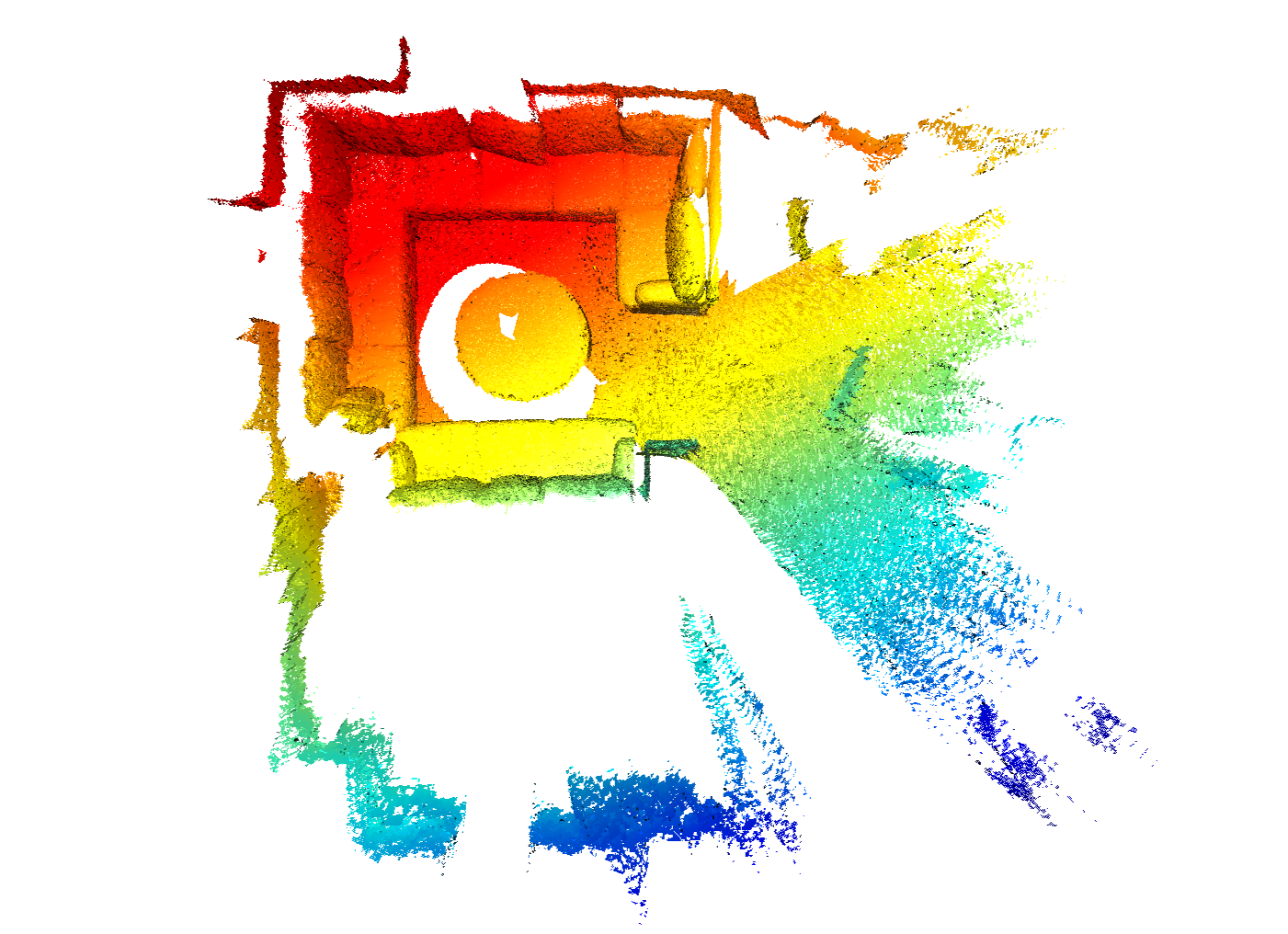}
    \end{minipage}}
    \caption{Qualitative comparison results.}
    \label{fig:vis}
\end{figure}

The quantitative results for 3DMatch, 3DLoMatch, and ScanNet are presented in Tab. \ref{tab:3dmatch}, Tab. \ref{tab:3dlomatch}, and Tab. \ref{tab:scannet}, respectively. Our method consistently demonstrates outstanding performance across all three benchmarks, irrespective of the pairwise registration methods used. For instance, our method achieves 97.1\% and 87.9\% on 3DMatch and 3DLoMatch when integrated with YOHO, surpassing the previous best by 0.9 and 5.6 percentage points, respectively. When adapted to ScanNet, our method continues to exhibit strong performance, further underscoring the effectiveness and generalizability of our design. Fig. \ref{fig:vis} provides some qualitative results.

\def\full{${}^{\heartsuit}$}
\def\prune{${}^{\clubsuit}$}
\def\sparse{${}^{\diamondsuit}$}

\begin{table}
    \caption{Registration recall on 3DMatch dataset. \full Full graph, \prune Pruned graph, \sparse Sparse graph.}
    \label{tab:3dmatch}
    \centering
    \resizebox{\linewidth}{!}{
    \begin{tabular}{lccccc|ccccc}
        \toprule
        & \multicolumn{5}{c}{FCGF} & \multicolumn{5}{c}{YOHO} \\
        \#Samples & 5000 & 2500 & 1000 & 500 & 250 & 5000 & 2500 & 1000 & 500 & 250 \\
        \midrule
        EIGSE3\full & 13.2 & 16.1 & 13.9 & 8.8 & 7.6 & 16.6 & 12.6 & 12.4 & 9.8 & 7.6 \\
        EIGSE3\prune & 60.3 & 62.5 & 66.5 & 68.5 & 73.9 & 43.6 & 45.7 & 50.3 & 49.9 & 53.7 \\
        EIGSE3\sparse & 45.9 & 46.8 & 40.1 & 37.7 & 35.3 & 58.9 & 54.5 & 39.2 & 36.6 & 25.4 \\
        IRLS\full & 47.0 & 44.4 & 47.8 & 44.3 & 35.5 & 41.9 & 41.3 & 45.8 & 45.5 & 33.6 \\
        IRLS\prune & 77.6 & 77.9 & 81.5 & 83.1 & 81.0 & 67.6 & 68.9 & 72.3 & 72.8 & 80.6 \\
        IRLS\sparse & 81.2 & 80.5 & 80.4 & 76.7 & 77.9 & 83.1 & 80.1 & 77.9 & 80.1 & 70.8 \\
        RotAvg\full & 62.0 & 65.3 & 69.8 & 59.7 & 61.0 & 66.6 & 65.6 & 68.3 & 60.8 & 56.5 \\
        RotAvg\prune & 86.5 & 84.8 & 84.6 & 86.9 & 82.5 & 77.3 & 78.0 & 79.0 & 79.2 & 80.9 \\
        RotAvg\sparse & 83.5 & 83.5 & 84.8 & 80.4 & 82.7 & 88.5 & 87.3 & 82.2 & 81.1 & 76.8 \\
        LITS\full & 72.8 & 77.7 & 76.8 & 73.0 & 67.5 & 78.4 & 68.1 & 77.3 & 71.2 & 67.2 \\
        LITS\prune & 82.3 & 83.2 & 83.8 & 85.2 & 78.7 & 82.5 & 82.5 & 86.5 & 83.6 & 83.4 \\
        LITS\sparse & 73.3 & 70.4 & 67.8 & 66.5 & 70.4 & 74.2 & 70.3 & 81.3 & 75.1 & 70.1 \\
        HARA\full & 83.5 & 83.7 & 85.2 & 79.8 & 77.8 & 78.8 & 79.1 & 80.8 & 79.6 & 76.7 \\
        HARA\prune & 88.2 & 91.7 & 89.7 & 88.9 & 89.1 & 83.6 & 84.7 & 89.7 & 88.9 & 85.7 \\
        HARA\sparse & 87.7 & 88.0 & 86.9 & 85.9 & 81.9 & 88.0 & 88.3 & 91.5 & 88.4 & 80.0 \\
        SGHR\full & 91.5 & 90.5 & 90.0 & 89.8 & 86.6 & 93.2 & 92.0 & 89.6 & \underline{92.8} & 80.0 \\
        SGHR\prune & \underline{93.9} & \underline{94.2} & \underline{91.8} & \underline{91.7} & \underline{91.8} & 95.2 & 93.8 & 92.7 & 89.9 & \underline{90.8} \\
        SGHR\sparse & 92.6 & 90.5 & 91.3 & 90.4 & 87.0 & \underline{96.2} & \underline{95.7} & \underline{95.3} & 92.2 & 90.7 \\
        Ours & \textbf{95.4} & \textbf{94.3} & \textbf{93.8} & \textbf{93.6} & \textbf{92.8} & \textbf{97.1} & \textbf{95.4} & \textbf{95.6} & \textbf{94.7} & \textbf{93.4} \\
        \bottomrule
    \end{tabular}
    }
\end{table}

\begin{table}
    \caption{Registration recall on 3DLoMatch dataset. \full Full graph, \prune Pruned graph, \sparse Sparse graph.}
    \label{tab:3dlomatch}
    \centering
    \resizebox{\linewidth}{!}{
    \begin{tabular}{lccccc|ccccc}
        \toprule
        & \multicolumn{5}{c}{FCGF} & \multicolumn{5}{c}{YOHO} \\
        \#Samples & 5000 & 2500 & 1000 & 500 & 250 & 5000 & 2500 & 1000 & 500 & 250 \\
        \midrule
        EIGSE3\full & 5.7 & 6.3 & 4.9 & 1.7 & 1.0 & 6.4 & 6.0 & 5.4 & 1.7 & 1.0 \\
        EIGSE3\prune & 43.0 & 45.5 & 47.6 & 51.8 & 51.5 & 33.8 & 35.0 & 35.7 & 34.4 & 37.2 \\
        EIGSE3\sparse & 29.8 & 30.3 & 25.5 & 21.7 & 20.9 & 39.8 & 36.9 & 26.1 & 24.5 & 13.0 \\
        IRLS\full & 35.7 & 30.7 & 32.9 & 25.4 & 20.1 & 25.9 & 27.2 & 28.0 & 27.0 & 17.4 \\
        IRLS\prune & 60.9 & 64.3 & 61.8 & 64.6 & 60.8 & 52.3 & 47.4 & 53.7 & 57.0 & 59.5 \\
        IRLS\sparse & 62.0 & 57.7 & 54.6 & 54.9 & 47.2 & 56.8 & 56.5 & 54.0 & 54.3 & 44.0 \\
        RotAvg\full & 47.0 & 49.8 & 54.1 & 43.0 & 42.4 & 45.0 & 46.0 & 46.1 & 45.8 & 37.2 \\
        RotAvg\prune & 71.4 & 72.8 & 71.8 & 73.4 & 64.8 & 58.6 & 59.2 & 62.7 & 64.9 & 64.2 \\
        RotAvg\sparse & 63.2 & 66.7 & 64.9 & 60.0 & 56.4 & 63.2 & 63.3 & 58.6 & 62.6 & 53.4 \\
        LITS\full & 58.9 & 61.0 & 59.1 & 52.5 & 46.2 & 62.8 & 51.3 & 59.9 & 50.7 & 44.8 \\
        LITS\prune & 67.3 & 68.8 & 68.3 & 66.5 & 64.2 & 66.0 & 68.0 & 68.3 & 66.2 & 62.7 \\
        LITS\sparse & 44.0 & 39.2 & 40.2 & 37.4 & 37.3 & 45.0 & 39.9 & 51.3 & 46.3 & 35.6 \\
        HARA\full & 64.5 & 67.3 & 65.4 & 66.7 & 59.8 & 60.5 & 60.7 & 65.7 & 62.0 & 60.6 \\
        HARA\prune & 75.9 & 80.4 & 76.8 & 74.3 & \underline{74.7} & 66.6 & 67.7 & 74.2 & 76.2 & 69.5 \\
        HARA\sparse & 68.8 & 71.3 & 70.6 & 66.5 & 55.9 & 68.9 & 68.8 & 74.1 & 68.0 & 56.6 \\
        SGHR\full & 80.6 & 79.5 & 75.5 & 77.0 & 69.1 & 76.8 & 77.5 & 74.4 & 73.0 & 63.6 \\
        SGHR\prune & \textbf{82.8} & \underline{81.1} & \underline{80.2} & \underline{79.5} & 73.1 & \underline{82.3} & 76.6 & 79.7 & \underline{76.1} & \underline{74.1} \\
        SGHR\sparse & 79.5 & 78.8 & 77.6 & 77.4 & 70.0 & 81.6 & \underline{80.5} & \underline{80.9} & 76.0 & 70.6 \\
        Ours & \underline{81.8} & \textbf{81.6} & \textbf{84.6} & \textbf{81.5} & \textbf{76.7} & \textbf{87.9} & \textbf{87.1} & \textbf{86.3} & \textbf{82.9} & \textbf{77.6} \\
        \bottomrule
    \end{tabular}
    }
\end{table}

\def\degree{${}^{\circ}$}
\begin{table}[t]
    \caption{Registration results on Scannet dataset with YOHO \#Samples=5000 setting. \full Full graph, \prune Pruned graph, \sparse Sparse graph.}
    \label{tab:scannet}
    \centering
    \resizebox{\linewidth}{!}{
    \begin{tabular}{lcccccc|cccccc}
        \toprule
        & \multicolumn{6}{c}{Rotation Error} & \multicolumn{6}{c}{Translation Error(m)} \\ 
        Method & 3\degree & 5\degree & 10\degree & 30\degree & 45\degree & Mean/Med & 0.05 & 0.1 & 0.25 & 0.5 & 0.75 & Mean/Med \\
        \midrule
        LMVR\full & 48.3 & 53.6 & 58.9 & 63.2 & 64.0 & 48.1\degree/33.7\degree & 34.5 & 49.1 & 58.5 & 61.6 & 63.9 & 0.83/0.55 \\
        EIGSE3\full & 19.7 & 24.4 & 32.3 & 49.3 & 56.9 & 53.6\degree/48.0\degree & 11.2 & 19.7 & 30.5 & 45.7 & 56.7 & 1.03/0.94 \\
        EIGSE3\prune & 40.8 & 46.3 & 51.9 & 61.2 & 65.7 & 40.6\degree/37.1\degree & 23.9 & 38.5 & 51.0 & 59.3 & 66.1 & 0.88/0.84 \\
        L1-IRLS\full & 38.1 & 44.2 & 48.8 & 55.7 & 56.5 & 53.9\degree/47.1\degree & 18.5 & 30.4 & 40.7 & 47.8 & 54.4 & 1.14/1.07 \\
        L1-IRLS\prune & 46.3 & 54.2 & 61.6 & 64.3 & 66.8 & 41.8\degree/34.0\degree & 24.1 & 38.5 & 48.3 & 55.6 & 60.9 & 1.05/1.01 \\
        RotAvg\full & 44.1 & 49.8 & 52.8 & 56.5 & 57.3 & 53.1\degree/44.0\degree & 28.2 & 40.8 & 48.6 & 51.9 & 56.1 & 1.13/1.05 \\
        RotAvg\prune & 50.2 & 60.1 & 65.3 & 66.8 & 68.8 & 38.5\degree/31.6\degree & 31.8 & 49.0 & 58.8 & 63.3 & 65.6 & 0.96/0.83 \\
        LITS\full & 52.8 & 67.1 & 74.9 & 77.9 & 79.5 & 26.8\degree/27.9\degree & 29.4 & 51.1 & 68.9 & 75.0 & 77.0 & 0.68/0.66 \\
        LITS\prune & 54.3 & 69.4 & 75.6 & 78.5 & 80.3 & 24.9\degree/19.9\degree & 31.4 & 54.4 & 72.3 & 76.7 & 79.6 & 0.65/0.56 \\
        HARA\full & 54.9 & 64.3 & 71.3 & 74.1 & 74.2 & 32.1\degree/29.2\degree & 35.8 & 54.4 & 66.3 & 69.7 & 72.9 & 0.87/0.75 \\
        HARA\prune & 55.7 & 63.7 & 69.0 & 70.8 & 72.1 & 34.7\degree/31.3\degree & 35.2 & 53.6 & 65.4 & 68.6 & 71.7 & 0.86/0.71 \\
        SGHR\full & 57.2 & 68.5 & 75.1 & 78.1 & 78.8 & 26.4\degree/19.5\degree & 39.4 & 61.5 & 72.0 & 75.2 & 77.6 & 0.70/0.59 \\
        SGHR\prune & \underline{59.4} & 71.9 & 80.0 & 82.1 & 82.6 & \underline{21.7}\degree/19.1\degree & \underline{39.9} & 63.0 & 74.3 & 77.6 & 80.2 & 0.64/\underline{0.47} \\
        SGHR\sparse & 59.1 & \underline{73.1} & \underline{80.8} & \underline{82.5} & \underline{83.0} & \underline{21.7\degree}/\underline{19.0\degree} & \underline{39.9} & \underline{64.1} & \underline{76.7} & \underline{79.0} & \underline{81.9} & \underline{0.56}/0.49 \\
        Ours & \textbf{59.9} & \textbf{74.0} & \textbf{85.1} & \textbf{87.9} & \textbf{88.5} & \textbf{17.8\degree}/\textbf{13.7\degree} & \textbf{41.8} & \textbf{63.1} & \textbf{79.3} & \textbf{83.1} & \textbf{85.5} & \textbf{0.50}/\textbf{0.43} \\
        \bottomrule
    \end{tabular}
    }
\end{table}

\subsection{Ablation Studies}
To verify the effectiveness of the proposed components, we conducted ablation studies on the 3D(Lo)Match dataset. Tab. \ref{tab:ablation} presents a summary of the ablation results. In Exp. (a), we tested the influence of coarse semantic selection by using only geometric clues to determine the growth ordering. In Exp. (b), we removed geometric reranking and relied solely on global features to determine the incremental sequence. Additionally, in Exp. (c), we used an overlap-based criterion from~\cite{guo2014accurate}. These experiments demonstrate the necessity of our two-stage design. In Exp. (d), we verified the effectiveness of our transformation averaging design. Finally, we conducted two experiments to validate the effectiveness of our meta-information update process. In Exp. (e), we simply merged newly registered frames by concatenation, while in Exp. (f), we calculated the mean of paired keypoints and descriptors instead of sampling.

\begin{table}[h]
    \caption{Ablation experiments. SS: Semantic stage. GS: Geometric stage. OR: Overlap-based retrieval. TR: Transformation refinement. RS: Reservoir sampling. MU: Mean update.  Tested with YOHO \#Samples=1000 setting.}
    \label{tab:ablation}
    \centering
    \begin{tabular}{c|ccccccc}
        \toprule
         & (a) w/o SS & (b) w/o GS & (c) w/ OR & (d) w/o TR & (e) w/o RS & (f) w/ MU & Full\\ 
        \midrule
        3DM   & 94.0 & 91.3 & 87.9 & 94.4 & 94.7 & 95.6 & 95.6\\
        3DLo  & 82.3 & 68.9 & 64.6 & 81.2 & 82.1 & 83.0 & 86.3\\
        \bottomrule
    \end{tabular}
\end{table}

\section{Conclusion}
In this paper, we propose a novel incremental multiview point cloud registration method. Our approach utilizes a two-stage coarse-to-fine point cloud retrieval process that combines the advantages of both semantic and geometric clues. Additionally, a single transformation averaging scheme and a Reservoir sampling-based update strategy further enhance robustness while reducing computational costs. Experiments on various datasets demonstrate the effectiveness of our proposed method.

\bibliography{ref}

\begin{thebibliography}{10}

\bibitem{huang2021predator}
Shengyu Huang, Zan Gojcic, Mikhail Usvyatsov, Andreas Wieser, and Konrad Schindler.
\newblock Predator: Registration of 3d point clouds with low overlap.
\newblock In {\em Proceedings of the IEEE/CVF Conference on computer vision and pattern recognition}, pages 4267--4276, 2021.

\bibitem{yu2021cofinet}
Hao Yu, Fu~Li, Mahdi Saleh, Benjamin Busam, and Slobodan Ilic.
\newblock Cofinet: Reliable coarse-to-fine correspondences for robust pointcloud registration.
\newblock {\em Advances in Neural Information Processing Systems}, 34:23872--23884, 2021.

\bibitem{qin2022geometric}
Zheng Qin, Hao Yu, Changjian Wang, Yulan Guo, Yuxing Peng, and Kai Xu.
\newblock Geometric transformer for fast and robust point cloud registration.
\newblock In {\em Proceedings of the IEEE/CVF conference on computer vision and pattern recognition}, pages 11143--11152, 2022.

\bibitem{yang2022one}
Fan Yang, Lin Guo, Zhi Chen, and Wenbing Tao.
\newblock One-inlier is first: Towards efficient position encoding for point cloud registration.
\newblock {\em Advances in Neural Information Processing Systems}, 35:6982--6995, 2022.

\bibitem{yu2023peal}
Junle Yu, Luwei Ren, Yu~Zhang, Wenhui Zhou, Lili Lin, and Guojun Dai.
\newblock Peal: Prior-embedded explicit attention learning for low-overlap point cloud registration.
\newblock In {\em Proceedings of the IEEE/CVF Conference on Computer Vision and Pattern Recognition}, pages 17702--17711, 2023.

\bibitem{jin2024multiway}
Shengze Jin, Iro Armeni, Marc Pollefeys, and Daniel Barath.
\newblock Multiway point cloud mosaicking with diffusion and global optimization.
\newblock {\em arXiv preprint arXiv:2404.00429}, 2024.

\bibitem{wang2023robust}
Haiping Wang, Yuan Liu, Zhen Dong, Yulan Guo, Yu-Shen Liu, Wenping Wang, and Bisheng Yang.
\newblock Robust multiview point cloud registration with reliable pose graph initialization and history reweighting.
\newblock In {\em Proceedings of the IEEE/CVF Conference on Computer Vision and Pattern Recognition}, pages 9506--9515, 2023.

\bibitem{huang2019learning}
Xiangru Huang, Zhenxiao Liang, Xiaowei Zhou, Yao Xie, Leonidas~J Guibas, and Qixing Huang.
\newblock Learning transformation synchronization.
\newblock In {\em Proceedings of the IEEE/CVF conference on computer vision and pattern recognition}, pages 8082--8091, 2019.

\bibitem{yew2021learning}
Zi~Jian Yew and Gim~Hee Lee.
\newblock Learning iterative robust transformation synchronization.
\newblock In {\em 2021 International Conference on 3D Vision (3DV)}, pages 1206--1215. IEEE, 2021.

\bibitem{boumal2014cramer}
Nicolas Boumal, Amit Singer, P-A Absil, and Vincent~D Blondel.
\newblock Cram{\'e}r--rao bounds for synchronization of rotations.
\newblock {\em Information and Inference: A Journal of the IMA}, 3(1):1--39, 2014.

\bibitem{gao2021incremental}
Xiang Gao, Lingjie Zhu, Zexiao Xie, Hongmin Liu, and Shuhan Shen.
\newblock Incremental rotation averaging.
\newblock {\em International Journal of Computer Vision}, 129:1202--1216, 2021.

\bibitem{guo2014accurate}
Yulan Guo, Ferdous Sohel, Mohammed Bennamoun, Jianwei Wan, and Min Lu.
\newblock An accurate and robust range image registration algorithm for 3d object modeling.
\newblock {\em ieee transactions on multimedia}, 16(5):1377--1390, 2014.

\bibitem{wu2023hierarchical}
Hao Wu, Li~Yan, Hong Xie, Pengcheng Wei, and Jicheng Dai.
\newblock A hierarchical multiview registration framework of tls point clouds based on loop constraint.
\newblock {\em ISPRS Journal of Photogrammetry and Remote Sensing}, 195:65--76, 2023.

\bibitem{shao2023global}
Shihao Shao, Kaifeng Chen, Arjun Karpur, Qinghua Cui, Andr{\'e} Araujo, and Bingyi Cao.
\newblock Global features are all you need for image retrieval and reranking.
\newblock In {\em Proceedings of the IEEE/CVF International Conference on Computer Vision}, pages 11036--11046, 2023.

\bibitem{sarlin2019coarse}
Paul-Edouard Sarlin, Cesar Cadena, Roland Siegwart, and Marcin Dymczyk.
\newblock From coarse to fine: Robust hierarchical localization at large scale.
\newblock In {\em Proceedings of the IEEE/CVF conference on computer vision and pattern recognition}, pages 12716--12725, 2019.

\bibitem{vitter1985random}
Jeffrey~S Vitter.
\newblock Random sampling with a reservoir.
\newblock {\em ACM Transactions on Mathematical Software (TOMS)}, 11(1):37--57, 1985.

\bibitem{zeng20173dmatch}
Andy Zeng, Shuran Song, Matthias Nie{\ss}ner, Matthew Fisher, Jianxiong Xiao, and Thomas Funkhouser.
\newblock 3dmatch: Learning local geometric descriptors from rgb-d reconstructions.
\newblock In {\em Proceedings of the IEEE conference on computer vision and pattern recognition}, pages 1802--1811, 2017.

\bibitem{dai2017scannet}
Angela Dai, Angel~X Chang, Manolis Savva, Maciej Halber, Thomas Funkhouser, and Matthias Nie{\ss}ner.
\newblock Scannet: Richly-annotated 3d reconstructions of indoor scenes.
\newblock In {\em Proceedings of the IEEE conference on computer vision and pattern recognition}, pages 5828--5839, 2017.

\bibitem{choy2019fully}
Christopher Choy, Jaesik Park, and Vladlen Koltun.
\newblock Fully convolutional geometric features.
\newblock In {\em Proceedings of the IEEE/CVF international conference on computer vision}, pages 8958--8966, 2019.

\bibitem{fischler1981random}
Martin~A Fischler and Robert~C Bolles.
\newblock Random sample consensus: a paradigm for model fitting with applications to image analysis and automated cartography.
\newblock {\em Communications of the ACM}, 24(6):381--395, 1981.

\bibitem{jiang2023robust}
Haobo Jiang, Zheng Dang, Zhen Wei, Jin Xie, Jian Yang, and Mathieu Salzmann.
\newblock Robust outlier rejection for 3d registration with variational bayes.
\newblock In {\em Proceedings of the IEEE/CVF conference on computer vision and pattern recognition}, pages 1148--1157, 2023.

\bibitem{zhang20233d}
Xiyu Zhang, Jiaqi Yang, Shikun Zhang, and Yanning Zhang.
\newblock 3d registration with maximal cliques.
\newblock In {\em Proceedings of the IEEE/CVF Conference on Computer Vision and Pattern Recognition}, pages 17745--17754, 2023.

\bibitem{hinterstoisser2016going}
Stefan Hinterstoisser, Vincent Lepetit, Naresh Rajkumar, and Kurt Konolige.
\newblock Going further with point pair features.
\newblock In {\em Computer Vision--ECCV 2016: 14th European Conference, Amsterdam, The Netherlands, October 11-14, 2016, Proceedings, Part III 14}, pages 834--848. Springer, 2016.

\bibitem{besl1992method}
Paul~J Besl and Neil~D McKay.
\newblock Method for registration of 3-d shapes.
\newblock In {\em Sensor fusion IV: control paradigms and data structures}, volume 1611, pages 586--606. Spie, 1992.

\bibitem{rusu2009fast}
Radu~Bogdan Rusu, Nico Blodow, and Michael Beetz.
\newblock Fast point feature histograms (fpfh) for 3d registration.
\newblock In {\em 2009 IEEE international conference on robotics and automation}, pages 3212--3217. IEEE, 2009.

\bibitem{wang2022you}
Haiping Wang, Yuan Liu, Zhen Dong, and Wenping Wang.
\newblock You only hypothesize once: Point cloud registration with rotation-equivariant descriptors.
\newblock In {\em Proceedings of the 30th ACM International Conference on Multimedia}, pages 1630--1641, 2022.

\bibitem{xu2022finet}
Hao Xu, Nianjin Ye, Guanghui Liu, Bing Zeng, and Shuaicheng Liu.
\newblock Finet: Dual branches feature interaction for partial-to-partial point cloud registration.
\newblock In {\em Proceedings of the AAAI Conference on Artificial Intelligence}, volume~36, pages 2848--2856, 2022.

\bibitem{jiang2024se}
Haobo Jiang, Mathieu Salzmann, Zheng Dang, Jin Xie, and Jian Yang.
\newblock Se (3) diffusion model-based point cloud registration for robust 6d object pose estimation.
\newblock {\em Advances in Neural Information Processing Systems}, 36, 2024.

\bibitem{arrigoni2016global}
Federica Arrigoni, Beatrice Rossi, and Andrea Fusiello.
\newblock Global registration of 3d point sets via lrs decomposition.
\newblock In {\em Computer Vision--ECCV 2016: 14th European Conference, Amsterdam, The Netherlands, October 11--14, 2016, Proceedings, Part IV 14}, pages 489--504. Springer, 2016.

\bibitem{arrigoni2018robust}
Federica Arrigoni, Beatrice Rossi, Pasqualina Fragneto, and Andrea Fusiello.
\newblock Robust synchronization in so (3) and se (3) via low-rank and sparse matrix decomposition.
\newblock {\em Computer Vision and Image Understanding}, 174:95--113, 2018.

\bibitem{gojcic2020learning}
Zan Gojcic, Caifa Zhou, Jan~D Wegner, Leonidas~J Guibas, and Tolga Birdal.
\newblock Learning multiview 3d point cloud registration.
\newblock In {\em Proceedings of the IEEE/CVF conference on computer vision and pattern recognition}, pages 1759--1769, 2020.

\bibitem{arrigoni2016spectral}
Federica Arrigoni, Beatrice Rossi, and Andrea Fusiello.
\newblock Spectral synchronization of multiple views in se (3).
\newblock {\em SIAM Journal on Imaging Sciences}, 9(4):1963--1990, 2016.

\bibitem{chatterjee2017robust}
Avishek Chatterjee and Venu~Madhav Govindu.
\newblock Robust relative rotation averaging.
\newblock {\em IEEE transactions on pattern analysis and machine intelligence}, 40(4):958--972, 2017.

\bibitem{li2022rago}
Heng Li, Zhaopeng Cui, Shuaicheng Liu, and Ping Tan.
\newblock Rago: Recurrent graph optimizer for multiple rotation averaging.
\newblock In {\em Proceedings of the IEEE/CVF Conference on Computer Vision and Pattern Recognition}, pages 15787--15796, 2022.

\bibitem{dong2018hierarchical}
Zhen Dong, Bisheng Yang, Fuxun Liang, Ronggang Huang, and Sebastian Scherer.
\newblock Hierarchical registration of unordered tls point clouds based on binary shape context descriptor.
\newblock {\em ISPRS Journal of Photogrammetry and Remote Sensing}, 144:61--79, 2018.

\bibitem{arandjelovic2016netvlad}
Relja Arandjelovic, Petr Gronat, Akihiko Torii, Tomas Pajdla, and Josef Sivic.
\newblock Netvlad: Cnn architecture for weakly supervised place recognition.
\newblock In {\em Proceedings of the IEEE conference on computer vision and pattern recognition}, pages 5297--5307, 2016.

\bibitem{radenovic2018fine}
Filip Radenovi{\'c}, Giorgos Tolias, and Ond{\v{r}}ej Chum.
\newblock Fine-tuning cnn image retrieval with no human annotation.
\newblock {\em IEEE transactions on pattern analysis and machine intelligence}, 41(7):1655--1668, 2018.

\bibitem{lee2020robust}
Seong~Hun Lee and Javier Civera.
\newblock Robust single rotation averaging.
\newblock {\em arXiv preprint arXiv:2004.00732}, 2020.

\bibitem{girshick2015fast}
Ross Girshick.
\newblock Fast r-cnn.
\newblock In {\em Proceedings of the IEEE international conference on computer vision}, pages 1440--1448, 2015.

\bibitem{lee2022hara}
Seong~Hun Lee and Javier Civera.
\newblock Hara: A hierarchical approach for robust rotation averaging.
\newblock In {\em Proceedings of the IEEE/CVF Conference on Computer Vision and Pattern Recognition}, pages 15777--15786, 2022.

\end{thebibliography}


\appendix

\section*{Appendix}

\section{Details about Single Rotation Averaging}
\label{app:averaging}
Denote the vectorization of an $n\times m$ matrix by $\text{vec}(\cdot)$ and its inverse by $\text{vec}^{-1}_{n\times m}(\cdot)$, the procedure of single rotation averaging is shown in Alg. \ref{alg:averaging}. 

\renewcommand{\algorithmicrequire}{\textbf{Input:}}  
\renewcommand{\algorithmicensure}{\textbf{Output:}} 
\begin{algorithm}[ht]
    \caption{Single rotation averaging}
    \begin{algorithmic}[1]
    \Require Rotation $\mathbf{R}$ from robust estimator and list of observed rotation with weight $\{\hat{\mathbf{R}}_i,w_i\}_{i=1}^M$.
    \Ensure Averaged rotation $\Bar{\mathbf{R}}$.

    \State $\mathbf{s}=\text{vec}(\mathbf{R})$
    
    \For{$it = 1,2,\cdots,10$} 
        \For{$i = 1,2,\cdots,M$}
            \State $v_i=\text{vec}(\mathbf{R}_i)-\mathbf{s}$
            \State $d_i=\Vert v_i\Vert$
        \EndFor
        \State $\mathbf{s}_{pre}=\mathbf{s}$
        \State $\mathbf{s}=\frac{\sum_{i=1}^Mw_iv_i/d_i}{\sum_{i=1}^Mw_i/d_i}$
        \If{$\Vert\mathbf{s}-\mathbf{s}_{pre}\Vert<0.001$}
            \State break
        \EndIf
    \EndFor
    \State $\Bar{\mathbf{R}}=\text{proj}_{SO(3)}(\text{vec}^{-1}_{3\times3}(\mathbf{s}))$
    
    \State \Return $\Bar{\mathbf{R}}$
    \end{algorithmic}
    \label{alg:averaging}
\end{algorithm}

\section{Probability Mass Function of the Bernoulli Distribution}
\label{app:pmf}
During the meta-information update, we use a multivariate Bernoulli distribution to address the redundancy issue, with the probability mass function of each variable $X_i$ specified. Here, we provide a more detailed derivation.
\begin{equation}
    \begin{aligned}
        P(X_i=k)&=\left(\frac{1}{r_i+1}\right)^k\left(1-\frac{1}{r_i+1}\right)^{1-k} \\
        &=\left(\frac{1}{r_i+1}\right)^k\left(\frac{r_i}{r_i+1}\right)^{1-k} \\
        &=\frac{r^{1-k}}{(r_i+1)^k(r_i+1)^{1-k}} \\
        &=\frac{r_i^{1-k}}{r_i+1},k\in\{0,1\},
    \end{aligned}
\end{equation}
where $r_i$ represents the redundancy count for point in pair $i$. As a result, for pair $i$, the possibility of selecting the new keypoint and descriptor is $\frac{1}{r_i+1}$, while the probability of keeping the existing ones is $\frac{r_i}{r_i+1}$.   

\section{Implementation Details}
We conducted experiments using PyTorch. For the global pooling layer, we utilized a NetVLAD with 64 clusters. Following the practice in~\cite{wang2023robust}, we train the model on an augmented 3DMatch training split. For each scene, we sampled alpha frames and for each frame, we randomly selected one keypoint, then picked beta nearest neighbor keypoints to yield varying overlap. We trained the model for 300 epochs using the Adam optimizer with an initial learning rate of $10^{-3}$ and a weight decay of $10^{-4}$, employing a cosine learning rate schedule. For the fusion operation in Eq. \ref{eq:fusion}, the cardinality of all $\mathcal{M}$ was set to 3. The value of $k$ in the first retrieval stage is set to 10, and the distance threshold $\tau$ is 0.07.

\section{Limitation}
The primary limitation of our method is its dependence on a target-agnostic descriptor-based pairwise registration method. Some recent pairwise methods use cross-attention to achieve information interaction, causing the extracted source features to vary with different target point clouds. This variability hampers our meta-shape construction and dynamic updates. Therefore, incorporating target-aware pairwise methods into our pipeline could be a promising direction for future research.

\end{document}